\documentclass[journal]{IEEEtran}

\usepackage{threeparttable}
\usepackage{tabularx}
\usepackage{booktabs}
\usepackage{verbatim}
\usepackage{amsopn}
\usepackage{amsmath}
\usepackage{multirow}
\usepackage{colortbl}
\usepackage{graphicx}

\usepackage[breaklinks=true,bookmarks=true]{hyperref}

\usepackage{amsfonts,amssymb}
\usepackage{bbm}
\usepackage{bm}

\usepackage{xspace}

\usepackage{color} 

\makeatletter
\DeclareRobustCommand\onedot{\futurelet\@let@token\@onedot}
\def\@onedot{\ifx\@let@token.\else.\null\fi\xspace}

\def\eg{\emph{e.g}\onedot} 
\def\ie{\emph{i.e}\onedot} 
 
\def\etc{\emph{etc}\onedot} \def\vs{\emph{vs}\onedot}
 
\def\etal{\emph{et al}\onedot}
\makeatother
\usepackage{epstopdf}

\ifCLASSINFOpdf
\else
\fi
\begin{document}
\title{
Dual-path TokenLearner for Remote Photoplethysmography-based Physiological Measurement with Facial Videos}
%

\author{Wei Qian, Dan Guo\textsuperscript{$\ast$}, Kun Li, Xilan Tian, Meng Wang\textsuperscript{$\ast$},~\IEEEmembership{Fellow,~IEEE}
\thanks{
This work was supported in part by the National Key R\&D Program of China (2022YFB4500600), and the National Natural Science Foundation of China under Grant 62272144, Grant 72188101, Grant 62020106007, and Grant U20A20183. \emph{(*Corresponding authors: Dan Guo and Meng Wang.)}} 
\thanks{
D. Guo and M. Wang are with Key Laboratory of Knowledge Engineering with Big Data (HFUT), Ministry of Education and School of Computer Science and Information Engineering, Hefei University of Technology (HFUT), Hefei, 230601, China, and are with Institute of Artificial Intelligence, Hefei Comprehensive National Science Center, Hefei, 230026, China (e-mail: guodan@hfut.edu.cn; eric.mengwang@gmail.com).

W. Qian and K. Li are with School of Computer Science and Information Engineering, Hefei University of Technology (HFUT), Hefei, 230601, China. (e-mail: qianwei.hfut@gmail.com; kunli.hfut@gmail.com).}
\thanks{X. Tian is with The 38th Research Institute of China Electronics Technology Group Corporation, Hefei, 230088, China (e-mail: xltian\_hf66@163.com).}
}

\markboth{IEEE TRANSACTIONS ON XXX, VOL. X, 2023}%
{Shell \MakeLowercase{\textit{et al.}}: Bare Demo of IEEEtran.cls for IEEE Journals}
%

\maketitle


\makeatletter 
\let\myorg@bibitem\bibitem
\def\bibitem#1#2\par{%
  \@ifundefined{bibitem@#1}{%
    \myorg@bibitem{#1}#2\par
  }{%
    \begingroup
      \color{\csname bibitem@#1\endcsname}%
      \myorg@bibitem{#1}#2\par
    \endgroup
  }%
}

\makeatother 

\begin{abstract}
Remote photoplethysmography (rPPG) based physiological measurement is an emerging yet crucial 
vision task, whose challenge lies in exploring accurate rPPG prediction from facial videos accompanied by noises of illumination variations, facial occlusions, head movements, \etc, in a non-contact manner. 
Existing mainstream CNN-based models make efforts to detect physiological signals by capturing subtle color changes in facial regions of interest (ROI) caused by heartbeats.
However, such models are constrained by the limited local spatial or temporal receptive fields in the neural units. 
Unlike them, a native Transformer-based framework called Dual-path TokenLearner (Dual-TL) is proposed in this paper, which utilizes the concept of learnable tokens to integrate both spatial and temporal informative contexts from the global perspective of the video. Specifically, the proposed Dual-TL uses a Spatial TokenLearner (S-TL) to explore associations in different facial ROIs, which promises the rPPG prediction far away from noisy ROI disturbances. 
Complementarily, a Temporal TokenLearner (T-TL) is designed to infer the quasi-periodic pattern of heartbeats, which eliminates temporal disturbances such as head movements.
The two TokenLearners, S-TL and T-TL, are executed in a dual-path mode. 
This enables the model to reduce noise disturbances for final rPPG signal prediction. 
Extensive experiments on four physiological measurement benchmark datasets are conducted. The Dual-TL achieves state-of-the-art performances in both intra- and cross-dataset testings, demonstrating its immense potential as a basic backbone for rPPG measurement. The source code is available at \href{https://github.com/VUT-HFUT/Dual-TL}{https://github.com/VUT-HFUT/Dual-TL}

\end{abstract}

\begin{IEEEkeywords}
Remote photoplethysmography, learnable token, physiological measurement, facial videos.
\end{IEEEkeywords}

\section{Introduction} 
\label{introduction}
\IEEEPARstart{P}{hysiological} signals such as heart rate (HR) and heart rate variability (HRV) 
are significant indicators of human health, which are widely used in the prevention and diagnosis of cardiovascular disease~\cite{schiweck2019heart,sunprivacy}. 
Two methodologies - electrocardiography (ECG) by using electrodes placed on the skin~\cite{gilgen2019rr} and photoplethysmography (PPG) by emitting and receiving LED light reflected from the skin with finger-clip contact sensors~\cite{pereira2020photoplethysmography} - are proposed to measure the physiological signals.
However, both methodologies require wearing a skin contact device. This is inconvenient and uncomfortable for participants, especially for those with sensitive skin such as newborns or burn patients. 
To overcome this limitation, the task of non-contact physiological measurement based on remote photoplethysmography (rPPG) is proposed and has attracted researchers' attention in recent years~\cite{de2013robust,tulyakov2016self,niu2019rhythmnet,kurihara2021non,li2018obf}. The rPPG-based measurement technique has already been promoted in applications such as driver monitoring~\cite{huang2020heart}, sleeping monitoring~\cite{van2020camera}, and face anti-spoofing~\cite{wang2022domain,yu2021deep}.

\begin{figure}[!t]
\centering
\includegraphics[width=1.0\linewidth]{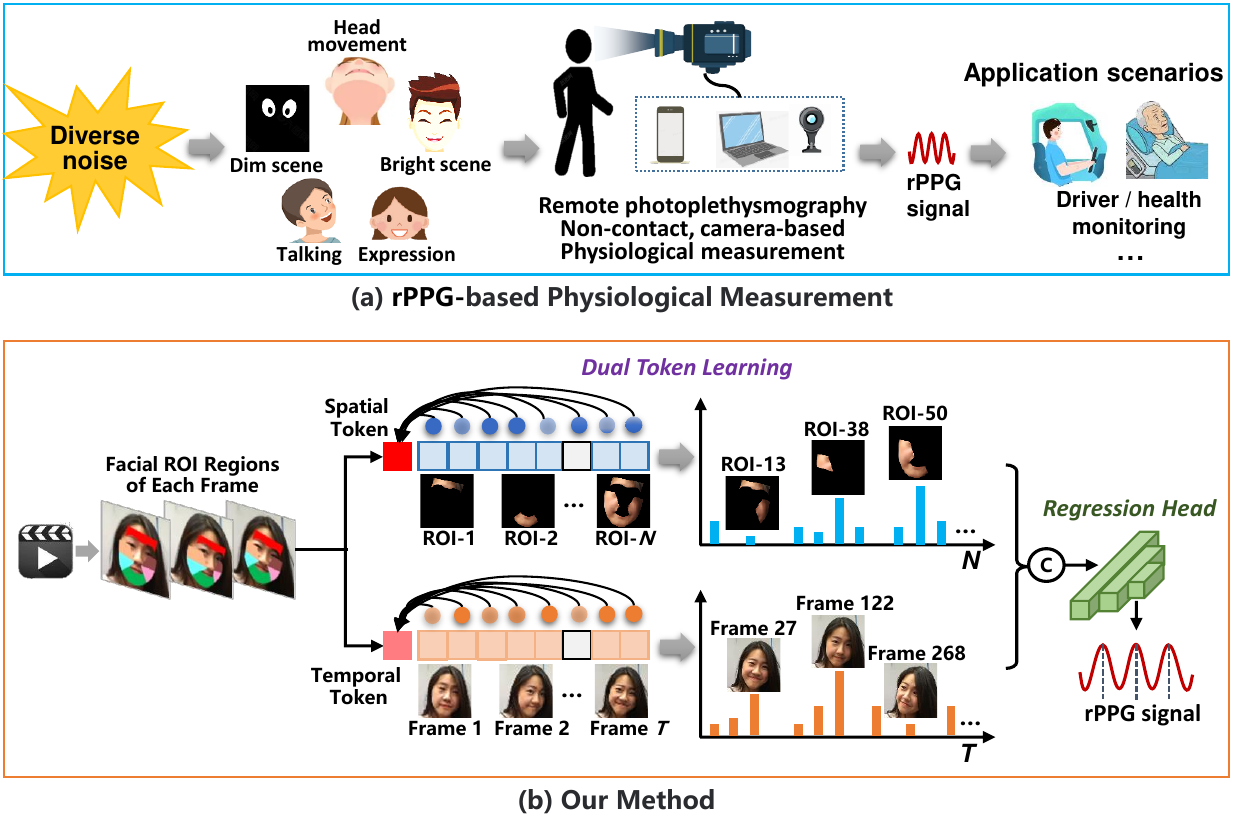}
\caption{(a) Illustration of the rPPG-based physiological measurement task. It faces diverse noisy disturbances. 
(b) Pipeline of our method for rPPG-based measurement. By imposing spatial and temporal token learning on the facial ROIs and video frames, respectively, we achieve a robust rPPG estimation.}
\label{fig:1}
\end{figure}

A fact is that rPPG-based physiological measurement~\cite{verkruysse2008remote,li2014remote,liu2021camera,yu2021facial,yu2021physformer,mcduff2020advancing} is a non-contact vision task. As an optical perception methodology, the rPPG recovers blood volume pulse (BVP) by observing subtle color changes on facial skin captured by a camera. Since the skin color changes are subtle, rPPG-based methods are inherently susceptible to environmental disturbances (\eg, illumination changes and diverse acquisition equipment) and subject movements (\eg, head movements, facial expressions, and partial occlusion). To mitigate these issues, several traditional approaches are proposed to disentangle the physiological signals from vision data, such as blind source separation~\cite{lewandowska2011measuring,lam2015robust,poh2010advancements} and color space transformation~\cite{de2013robust,wang2016algorithmic}. 
But yet, they are limited by some prior knowledge or assumptions such as skin reflection model~\cite{de2013robust,wang2016algorithmic} or linear combination assumption~\cite{poh2010advancements,poh2010non}. 
These methods are not equipped to handle sophisticated noises well and bring significant performance degradation.

Recently, with the great success of deep learning in computer vision~\cite{liproposal,guodadnet,guo2019hierarchical}, several deep learning-based methods~\cite{chen2018deepphys,vspetlik2018visual,lee2020meta,lu2021dual,niu2020video,gideon2021way,wang2022synthetic} have been developed. 
Early methods apply hand-crafted representations (\eg, time-frequency map~\cite{hsu2017deep}, spatial-temporal map~\cite{niu2018vipl}) for rPPG prediction. 
Subsequently, Liu~\etal~\cite{liu2020multi} performs the CNN on the difference of consecutive frames to extract fine-grained visual representation. 
Diverse attention mechanisms~\cite{chen2018deepphys,niu2019robust} are proposed to enhance the representation of salient visual variation. 
Moreover, to refine the representation learning, some methods 
propose complicated strategies to 
distinguish real physiological signals from the noises, such as cross-verified feature disentangling~\cite{niu2020video} or Dual-GAN~\cite{lu2021dual}.
By considering the advancement in multiple fields and potential beyond the CNN~\cite{vaswani2017attention, dosovitskiy2020image}, Transformer is a desirable candidate solution to address the rPPG measurement task. 
There are merely two Transformer-based works for this task~\cite{yu2021physformer,liu2021efficientphys}, which consistently focus on long-range dependencies in the video for representation learning. In this work, we utilize two learnable tokens rather than feature representation in the Transformer framework to retain the informative contexts for rPPG-based measurement.

As shown in Fig.~\ref{fig:1} (a), face regions are exposed to different noise disturbances, resulting in irregularly captured signals. In this study, we aim to exploit the spatial and temporal clues associated with rPPG signals in facial videos. 
To be specific, we propose a native framework for the video-based rPPG measurement task, namely 
an elegant and neat Transformer-based architecture named Dual-path TokenLearner (Dual-TL).
The pipeline of our method is illustrated in Fig.~\ref{fig:1} (b), in which we decouple both spatial and temporal token learning.  
We design a Spatial TokenLearner (S-TL) module to learn the association among the different facial ROI combinations, which adaptively responds to appropriate facial regions. 
Complementarily, a Temporal TokenLearner (T-TL) is designed to learn the quasi-periodic pattern of heartbeats, which is expected to eliminate temporal disturbances such as head movements. This work is inspired by the benefits of token learning~\cite{dosovitskiy2020image,carionend}. Based on the learnable token's merits as below, we utilize it to explore the global clues in videos. 1) Token globally aggregates the informative contexts in the video from a spatial perspective (across different facial ROI combinations) and temporal perspective (across consecutive frames); 2) Token is independent of video data. It is randomly initialized and is learned by model optimization. It avoids the data bias towards the noises of race, illumination, movement, acquisition device, \etc.

After token learning, the learned spatial token and temporal token are concatenated and input into an rPPG regression head module named FFN$_{reg}$. Finally, the FFN$_{reg}$ outputs an rPPG signal sequence. We experiment on four prevalent physiological measurement datasets, including UBFC-rPPG~\cite{bobbia2019unsupervised}, PURE~\cite{stricker2014non}, MMSE-HR~\cite{tulyakov2016self}, VIPL-HR~\cite{niu2019rhythmnet}. The proposed Dual-TL obtains a series of state-of-the-art records. 
Remarkably, our method achieves improvements of 38.8\%, 43.8\%, and 9.9\% in RMSE over the strongest competitors on the intra-dataset testing of UBFC-rPPG, PURE, and VIPL-HR datasets, respectively. Besides, the cross-dataset testing result of our method on the MMSE-HR dataset improves the RMSE by 4.7\%. 
Furthermore, extensive ablation studies and quantitative analysis are conducted to demonstrate that our Dual-TL is interpretable and effective for rPPG measurement.

In summary, we make three-fold contributions as follows:
\begin{itemize}
    \item We design a simple yet effective Transformer-based Dual-path TokenLearner (\ie, Dual-TL) consisting of a {Spatial TokenLeaner} (S-TL) and a {Temporal TokenLeaner} (T-TL), in which two learnable tokens are designed to retain and purify the informative spatial and temporal contexts in facial videos for rPPG measurement. 
    \item By implementing S-TL and T-TL in parallel, the distinct Dual-TL fully exploits complementary spatiotemporal dependencies in facial videos to eliminate external disturbances (\eg, environmental conditions and individual characteristics). S-TL and T-TL are performed to interact with different facial ROI combinations and consecutive frames, respectively. 
    \item Our model achieves state-of-the-art performance on four available benchmark datasets for remote physiological measurement, outperforming the best-published results by a considerable margin. Extensive ablation studies and visualization results show the effectiveness of Dual-TL in tackling complicated noise scenarios including single-source noise and multi-source noise cases.
\end{itemize}

The remainder of this paper is organized as follows. Section~\ref{related work} briefly reviews the related work. In Section~\ref{method}, we elaborate on the proposed Dual-TL. Section~\ref{experiments} analyzes experimental results on four physiological benchmark datasets to demonstrate the superiority of Dual-TL. In Section~\ref{conclusions}, we summarize the work with some concluding remarks.

\section{Related Work} \label{related work}
Physiologists discovered that optical absorption on human skin varies periodically along with the changeable BVP signals, which indicate the heartbeat pattern~\cite{wieringa2005contactless}. 
Motivated by this, Verkruysse \etal~\cite{verkruysse2008remote} first proposed the rPPG measurement task in a non-contact manner of facial video captured by a commodity camera. Existing work can be divided into two categories -- traditional hand-crafted-based methods~\cite{de2013robust,tulyakov2016self,li2014remote,lewandowska2011measuring,lam2015robust,poh2010advancements,wang2016algorithmic,poh2010non,wang2014exploiting} and deep learning-based methods~\cite{yu2021physformer,chen2018deepphys,vspetlik2018visual,lee2020meta,lu2021dual,niu2020video,gideon2021way,liu2021efficientphys,yu2019remote1,nowara2021benefit}. 

\subsection{Vision-based Physiological Measurement}
\subsubsection{\textbf{Hand-crafted-based methods}}
Early studies made efforts to improve the signal-to-noise rate of vision data, referring to blind source separation (BSS) and color space transformation approaches. BSS approaches such as principal component analysis (PCA)~\cite{lewandowska2011measuring} and independent component analysis (ICA)~\cite{lam2015robust,poh2010advancements,poh2010non} were used to disentangle rPPG signal from the noises (\eg, head movement or illumination variation). 
For color space transformation, the green (G) channel was demonstrated that strongly reflected the rPPG signal among RGB channels~\cite{verkruysse2008remote,li2014remote}. Haan \etal ~\cite{de2013robust} proposed a chrominance-based color space projection (CHROM) algorithm, which realized a linear combination of RGB channels for improving motion robustness. Wang \etal~\cite{wang2016algorithmic} combined both optical and physiological properties of skin and designed a plane-orthogonal-to-skin (POS) algorithm. 
Besides, to further capture the skin-reflection changes to the human face, some methods~\cite{tulyakov2016self,lewandowska2011measuring,lam2015robust,poh2010advancements,wang2016algorithmic} explored facial regions of interest (ROI) in the video.
Different face landmark-based ROI approaches were designed to obtain high-quality facial regions or reduce background interference~\cite{tulyakov2016self,niu2017continuous}, such as skin segmentation-based ROI~\cite{de2013robust,li2014remote,wang2014exploiting}, rectangular ROIs located on the center of face ~\cite{poh2010non} and the bottom of face~\cite{li2014remote}. Based on the landmark points, Niu~\etal proposed a spatial-temporal map (STmap)~\cite{niu2018vipl} and an advanced version (multi-scale STmap named MSTmap)~\cite{niu2020video}; both methods specialized in retrieving the spatial characteristics from the partial and whole facial areas.

Admittedly, preliminary hand-crafted methods have promoted the development of rPPG physiological measurement. However, they heavily rely on prior knowledge or certain skin or color space assumptions~\cite{de2013robust,wang2016algorithmic, poh2010non} and require stationary subjects without head movement and 
illumination variation. To summarize, traditional methods are still sensitive to less-constrained situations. 
 
\subsubsection{\textbf{Deep learning-based methods}}
With the advancement of deep learning in the computer vision field~\cite{guohierarchical,guoconnectionist}, several networks have been proposed for the rPPG measurement task~\cite{yu2021physformer,chen2018deepphys,vspetlik2018visual,lee2020meta,lu2021dual,niu2020video,gideon2021way}. Hsu \etal ~\cite{hsu2017deep} encoded each video into a two-dimensional time-frequency representation and cascaded it with the famous VGG15~\cite{simonyan2014very} for rPPG prediction. 
{\v{S}}petl{\'\i}k~\etal~\cite{vspetlik2018visual} proposed a two-step CNN backbone with a novel end-to-end alternating training optimization.  
Chen \etal ~\cite{chen2018deepphys} proposed an attention-based CNN named DeepPhys by leveraging a skin reflex-based motion representation to eliminate the negative effect of head movements. 
But these CNN-based methods primarily focus on spatial learning.
To further capture the temporal contexts, some spatiotemporal modeling networks are proposed. 
For example, RhythmNet~\cite{niu2019rhythmnet} and PhysNet~\cite{yu2019remote1} leveraged CNN for spatial learning and RNN for temporal modeling. However, RNN always exhibits gradient attenuation across overlong temporal spans.
3D CNN-based methods~\cite{yu2019remote1,tsou2020siamese} are also proposed to perform spatio-temporal learning with 3D convolutional kernels.  
Besides, Liu~\etal~\cite{liu2020multi} proposed a novel multi-task temporal shift convolutional attention network named MTTS-CAN, which adopted a tensor-shift module instead of 3D CNN to achieve efficient spatiotemporal modeling. 
Anyway, they face a common drawback of the local receptive field.

In addition, some novel approaches distinguished the rPPG signal from noises, such as cross-verified disentangling~\cite{niu2020video} and Generative Adversarial Networks (GANs)~\cite{lu2021dual,song2021pulsegan}.
In Dual-GAN~\cite{lu2021dual}, Lu~\etal constructed both a noise and a noise-free generator to generate a synthetic STmap similar to the STmap obtained from real videos. 
Its novelty is to enhance the robustness of rPPG predictor against unseen noises by online data augmentation. 
In PulseGAN~\cite{song2021pulsegan}, a generator transfers a rough pulse signal derived from CHROM to a target rPPG signal, while a discriminator distinguishes the rPPG signal from the reference PPG signal. 
Moreover, some researchers have explored self-supervised methods such as~\cite{gideon2021way,suncontrast,luneuron}.  These self-supervised approaches focus on contrastive learning between samples and have shown promising performance in rPPG tasks. In this paper, we mainly focus on the supervised approach to address the challenges of remote physiological measurement.

\subsection{Transformer for Vision-based Physiological Measurement}
To our knowledge, there are merely two Transformer-based works~\cite{yu2021physformer,liu2021efficientphys} for rPPG-based physiological measurement.
Liu~\etal~\cite{liu2021efficientphys} proposed two Transformer-based networks named EfficientPhys-T1 and EfficientPhys-T2. EfficientPhys-T1 utilized the merit of the tensor-shift module (TSM)~\cite{lin2019tsm} for spatiotemporal learning and embedded the TSM into the famous Swin-Transformer~\cite{liu2021swin}, while EfficientPhys-T2 is a lightweight version of EfficientPhys-T1 for on-device inference. 
Besides, PhysFormer~\cite{yu2021physformer} is an end-to-end video Transformer, which adaptively aggregates both local and global spatiotemporal features. The technique contribution of PhysFormer is the temporal difference Transformer, which addresses long-range spatiotemporal interaction for quasi-periodic rPPG modeling. 
Another somewhat related work is TransRPPG~\cite{yu2021transrppg} which addressed the attack detection task with an assumption that rPPG signal can only be observed on genuine faces rather than 3D masked faces. 
Essentially, existing Transformer-based methods made efforts to achieve better spatio-temporal representation of the video for rPPG prediction, while we use two newly added token vectors to aggregate the spatio-temporal contexts from the video for rPPG prediction.

\begin{figure*}[t]
\centering
\includegraphics[width=1.0\linewidth]{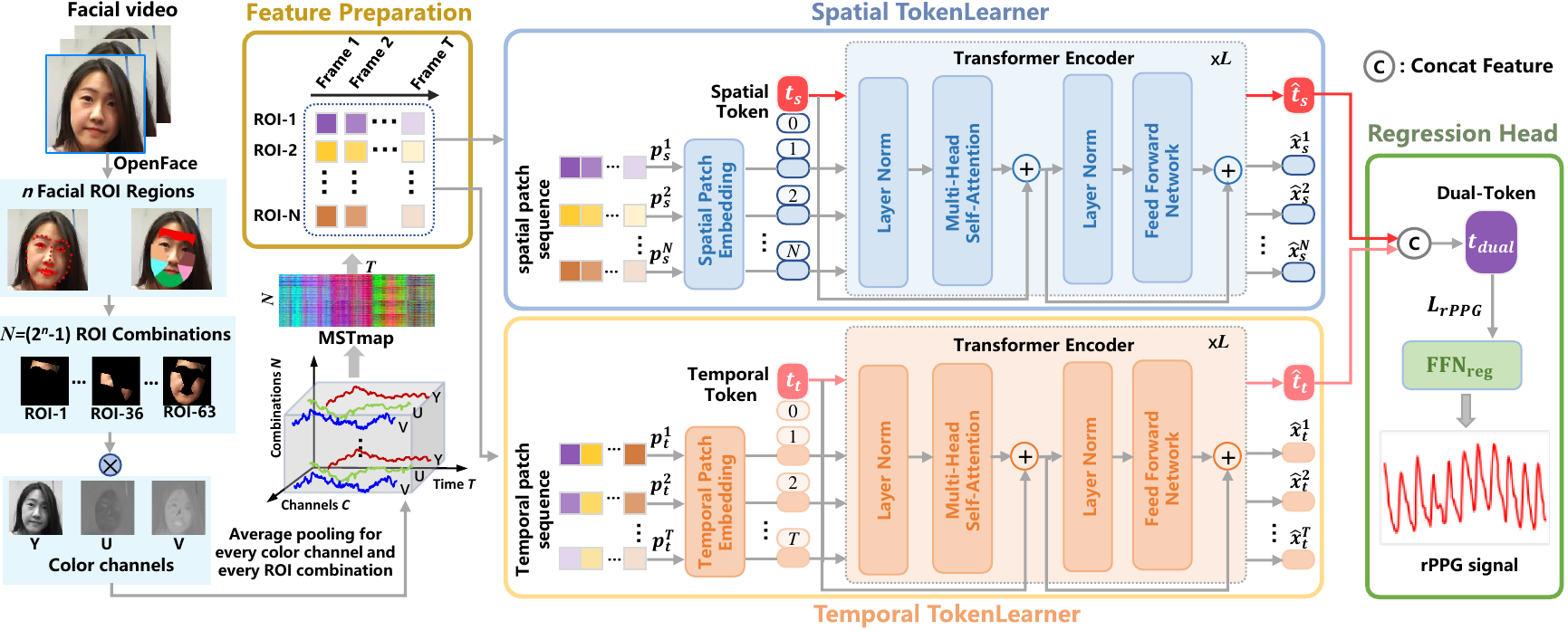}
\caption{{\textbf{Architecture overview of the proposed Dual-path TokenLearner (Dual-TL).} First, an input facial video with $T$ frames is processed to the MSTmap representation ${\bf M} \in \mathbb{R}^{N \times C \times T}$, where $N$ is the number of landmark-based facial ROI combinations. Subsequently, the MSTmap is split into two types of patch sequences, \ie, a $N$-length spatial patch sequence ${\bf X}_s=\{{\bf p}^{1}_{s},{\bf p}^{2}_{s},{\bf p}^{3}_{s},\cdots,{\bf p}^{N}_{s}\}$ and a $T$-length temporal patch sequence ${\bf X}_t= \{{\bf p}^{1}_{t},{\bf p}^{2}_{t},{\bf p}^{3}_{t},\cdots,{\bf p}^{T}_{t}\}$, where ${\bf X}_s$ describes the MSTmap at facial ROI combination level and ${\bf X}_t$ describes the MSTmap at the frame level. In the Transformer-based Dual-TL, a Spatial TokenLearner (\textbf{S-TL}) leverages a learnable token ${\bf\hat{t}}_s$ to perform the spatial context learning in the facial ROI sequence. Complementarily, a temporal TokenLearner (\textbf{T-TL}) is designed for exploring temporal contextual representation by using a learnable token ${\bf\hat{t}}_t$. Finally, the dual-token ${\bf t}_{dual}$ (by concatenating ${\bf\hat{t}}_s$ and ${\bf\hat{t}}_t$) is input into a regression head to predict rPPG. Note that regression head is trained with a loss function $\mathcal{\bm {L}}_{rPPG}$.}}
\label{fig:pipeline}
\end{figure*}

\section{Proposed Method} \label{method}
In this paper, we devote to rPPG-based physiological signal estimation from facial videos in a non-contact manner. We propose a native Dual-path TokenLearner method called \textbf{Dual-TL}. In \textbf{Dual-TL}, we design a spatial and a temporal token to separately learn the associations in landmark-based facial ROI combinations and in consecutive frames. We introduce the methodological details below. 

\subsection{Method Overview \& Feature Preparation}
\subsubsection{Method Overview} As shown in Fig.~\ref{fig:pipeline}, the Dual-TL consists of four modules, namely Feature Preparation, Spatial TokenLearner, Temporal TokenLearner, and Regression Head modules. 
At first, we detect facial ROIs and build a 2D multi-scale spatial-temporal map (MSTmap, detailed in Section~\ref{MSTmap construction}). Then, we feed the MSTmap into the dual path of spatial and temporal learning in parallel (detailed in Section~\ref{Dual-path TokenLearner}). Specifically, we design a Spatial TokenLearner (\textbf{S-TL}) and a Temporal TokenLearner (\textbf{T-TL}) based on the Transformer-Encoder for rPPG prediction.
\textbf{S-TL} models the spatial discrepancies of face ROI combinations, while \textbf{T-TL} emphasizes the temporal variation along the timeline. 
Next, we utilize these two learnable tokens to output the predicted rPPG signals by an rPPG regression head. Finally, a loss function based on Negative Pearson Correlation is applied for model optimization (detailed in Section~\ref{Traininf}).

\subsubsection{MSTmap Preparation} \label{MSTmap construction}
In this work, we adopt the multi-scale spatiotemporal map (MSTmap)~\cite{niu2020video} for feature preparation, which considers all possible facial ROI combinations.
For each video frame, we detect the facial landmarks by using OpenFace~\cite{baltrusaitis2018openface}, thus obtaining $n$ = 6 facial ROIs, namely the forehead, left upper cheek, left lower cheek, right upper cheek, right lower, and chin. 
As shown in Fig.~\ref{fig:pipeline}, we get a ROI combination set $R = \{R_{1}, R_{2}, \cdots, R_{N}\}$, where $N$ = ($2^n$ -1). For each ROI combination, we average the pixel values under $\{R, G, B\}$ or $\{Y, U, V\}$ channels. 
Therefore, given a facial video segment, we obtain the MSTmap representation $\bf{M}$ with the size of $N \times C\times T$, where $N$ is the size of the ROI combination set, $C$ is the channel number of color space, and $T$ is the frame number of the video segment. Up to now, we obtain a preliminary representation of each facial video. 

\subsection{Dual-path TokenLearner} \label{Dual-path TokenLearner}
To explore spatial and temporal associations in the facial video, we propose a Transformer-based TokenLearner framework with a self-attention mechanism. 
Concretely, we built a dual-path model with complementary Spatial and Temporal TokenLearners (\ie, \textbf{S-TL} and \textbf{T-TL}). Then, we further concatenate the learnable tokens (\ie, ${\bf{t}}_s$ and ${\bf{t}}_t$) into a fusion representation ${\bf t}_{dual}$. 
By exploiting spatial and temporal TokenLearners based on Transformer, \textbf{S-TL} path explores associations of different ROI combinations over the whole face, which can discover the optical absorption in the stable appearances. 
Complementarily, \textbf{T-TL} path models the temporal variation of facial regions along the timeline. To summarize, the model is required to integrate spatial and temporal informative contexts thereby suppressing noises of lighting conditions, hair occlusion, head movements, \etc.

\subsubsection{\textbf{Spatial TokenLearner}} 
We design a learnable token ${\bf t}_s\in \mathbb{R}^{D}$ and propose a concise \textbf{Spatial TokenLearner (S-TL)} with Transformer-Encoder~\cite{dosovitskiy2020image}. 
Given a MSTmap ${\bf M} \in \mathbb{R}^{N \times C \times T}$, we convert it into a flattened non-overlapping sequence ${\bf X}_s=\{{\bf p}^{1}_{s},{\bf p}^{2}_{s},{\bf p}^{3}_{s},\cdots,{\bf p}^{N}_{s}\}$, 
where ${\bf p}^{i}_{s}\in \mathbb{R}^{1 \times (CT)}$ ($i \in [1,N]$)  indicates the representation of $i$-th ROI combination (\ie, a spatial patch). 
Then, we jointly consider the spatial patch sequence ${\bf X}_s$ and the spatial token ${\bf t}_s$. Moreover, we perform a \textbf{Spatial Patch Embedding} ({\bf SPE}) operation to introduce a positional embedding ${\bf f}^{s}_{pos}$ as the auxiliary input of ${\bf X}_s$ and ${\bf t}_s$. 
The {\bf SPE} process is formulated as follows:
\begin{equation}
\label{eq:s-spe}
\begin{aligned}
	&{\bf z}^{s}_{0} = {\bf SPE}({\bf X}_s, {\bf t}_s) \Leftrightarrow\\
	&\left\{\begin{array}{l}
	    {\bf X}^{'}_{s}= [{\bf p}^{1}_{s}{\bf E}_{s};{\bf p}^{2}_{s} {\bf E}_{s};\cdots;{\bf p}^{N}_{s}{\bf E}_{s}] \in \mathbb{R}^{N\times D} ,\\
	    {\bf z}^{s}_{0}= [{\bf t}_{s};{\bf X}^{'}_{s}]+{\bf f}^{s}_{pos}\in \mathbb{R}^{(N+1)\times D},
	\end{array}\right.
\end{aligned}
\end{equation}
where ${\bf E}_s \in \mathbb{R}^{(CT) \times D}$ denotes a linear projection layer to map each ROI combination patch into a $D$-dim vector. The variable ${\bf z}^s_0$ indicates that we concatenate the token ${\bf t}_s$ and the spatial feature embedding ${\bf X}^{'}_{s}$, and introduce the positional embedding ${\bf f}^{s}_{pos}$.

After the above-mentioned SPE embedding, we perform the Transformer-Encoder~\cite{dosovitskiy2020image} to learn the association of spatial patches. The architecture of \textbf{S-TL} is simple. The spatial token is used to retain the spatial context for rPPG prediction. Specifically, the Transformer-Encoder module includes a stack of $L$ encoder layers. At the $\ell$-th layer, we use the multi-head self-attention (\textbf{MHSA}), feedforward network (\textbf{FFN}), and Layer Norm (\textbf{LN}) operations to perform the spatial interaction between the token and different ROI combination patches, 
which is formulated as follows:
\begin{align}
\label{eq:s-trans-spe}
&\left\{\begin{array}{l}
    {\bf z'}^{s}_{\ell-1} = \bf{MHSA}(\bf{LN}({\bf z}^{s}_{\ell-1}))+{\bf z}^{s}_{\ell-1}, \\
    {\bf z}^{s}_{\ell} = {\bf FFN}({\bf LN}({\bf z'}^s_{\ell-1}))+{\bf z'}^s_{\ell-1}, \ell \in [1,L].
\end{array}\right.
\end{align}

To summarize, the \textbf{S-TL} module 
is formulated as follows:
\begin{equation}
\begin{aligned}
	&{\bf \hat t}_s = {\bf S\text{-}TL}({\bf X}_s, {\bf t}_s) \Leftrightarrow\\
	&\left\{\begin{array}{l}
	    {\bf z}^{s}_{0}= {\bf SPE}({\bf X}_s, {\bf t}_s) ,\\
	    {[{\bf\hat t}_s, {\bf\hat X}_s]} = {\bf Transformer\text{-}Encoder} ({\bf z}^s_0).
	\end{array}\right.
\end{aligned}
\label{eq3}
\end{equation}

\subsubsection{\textbf{Temporal TokenLearner}}
To learn temporal relationships in the facial video, we design a \textbf{Temporal TokenLearner} ({\bf T-TL}) module.
Here, we flatten the original MSTmap $\bm M$ into ${\bf X}_t=\{{\bf p}^{1}_{t},{\bf p}^{2}_{t},{\bf p}^{3}_{t},\cdots,{\bf p}^{T}_{t}\}$, where ${\bf p}^{j}_{t}\in \mathbb{R}^{(N C) \times 1} (j \in [1,T])$ indicates the flattened MSTmap vector at the $j$-th timestamp (frame).
Next, we perform the {\bf T-TL} to model the interaction among the temporal token ${\bf t}_t$ and the temporal feature embeddings ${\bf X}^{'}_t$.
As the same to the spatial token ${\bf t}_s$, 
we realize a \textbf{Temporal Patch Embedding (TPE)} as below:
\begin{equation}
\label{eq:t-tpe}
\begin{aligned}
	&{\bf z}^{t}_{0} = {\bf TPE}({\bf X}_t, {\bf t}_t) \Leftrightarrow\\
	&\left\{\begin{array}{l}
	    {\bf X}^{'}_{t}= [{\bf p}^{1}_{t}{\bf E}_{t};{\bf p}^{2}_{t} {\bf E}_{t};\cdots;{\bf p}^{T}_{t}{\bf E}_{t}] \in \mathbb{R}^{T\times D} ,\\
	    {\bf z}^{t}_{0}= [{\bf t}_{t};{\bf X}^{'}_{t}]+{\bf f}^{t}_{pos}\in \mathbb{R}^{(T+1)\times D},
	\end{array}\right.
\end{aligned}
\end{equation}
where ${\bf f}^{t}_{pos} \in \mathbb{R}^{(T+1) \times D}$ denotes the positional embedding in the temporal token learning, and ${\bf E}_t \in \mathbb{R}^{(N C) \times D}$ denotes a linear projection layer to map the MSTmap at each timestamp into a $D$-dim vector.

To summarize, different from the \textbf{S-TL} focusing on mining the influence of each ROI combination patch, the \textbf{T-TL} devotes to learning the periodic pattern along the timeline. Thus, this Temporal TokenLearner is formulated as follows:
\begin{equation}
\begin{aligned}
	&{\bf \hat t}_t = {\bf T\text{-}TL}({\bf X}_t, {\bf t}_t) \Leftrightarrow\\
	&\left\{\begin{array}{l}
	    {\bf z}^{t}_{0}= {\bf TPE}({\bf X}_t, {\bf t}_t) ,\\
	    {[{\bf\hat t}_t, {\bf\hat X}_t]} = {\bf Transformer\text{-}Encoder} ({\bf z}^t_0).
	\end{array}\right.
\end{aligned}
\label{eq5}
\end{equation}

\subsubsection{Dual Spatial-Temporal Path} 
Merely leveraging a single path from the spatial or temporal perspective is insufficient to discover the rPPG periodic pattern. As the above-mentioned, ${\bf\hat t}_s$ is used to capture the spatial association of ROI combinations, whereas ${\bf\hat t}_t$ is used to learn the temporally facial variation in the video. Thus, we integrate both tokens ${\bf\hat t}_s$ and ${\bf\hat t}_t$ to better address the rPPG prediction. This entire operation of Section \ref{Dual-path TokenLearner} is defined as \textbf{Dual-path TokenLearner (Dual-TL)}.
Specifically, given the MSTmap sequence of each video, we reshape it into ${\bf X}_s\in \mathbb{R}^{(CT)\times N}$ and ${\bf X}_t\in \mathbb{R}^{(CN)\times T}$ as the input features. The dual-token is learned by ${\bf t}_{dual}$ as formulated below:
\begin{equation}
	{\bf t}_{dual} = {\bf Dual\text{-}TL}({\bf X}_{s}, {\bf X}_{t}, {\bf t}_{s}, {\bf t}_{t})	= [{\bf\hat t}_{s}; {\bf\hat t}_{t}],
 \label{eq6}
\end{equation}
where [;] denotes the concatenation operation.

\subsection{Training} \label{Traininf}
In this subsection, our goal is to predict a sequential rPPG signal $s_{pre}$ as consistent as the ground-truth label (\eg, BVP or PPG signal captured by PPG methodologies). 

\subsubsection{rPPG Regression Head} 
This head consists of two fully-connected layers and a GELU activation layer, denoted as $\bf{FFN_{reg}}(\cdot)$. The token ${\bf t}_{dual}$ is fed into $\bf{FFN_{reg}}$ and the output rPPG signal $s_{pre}$ is calculated as follows:
\begin{align}
& s_{pre}={\bf FFN_{reg}}({\bf t}_{dual})=FC_{2}({\bf GELU}(FC_{1}({\bf t}_{dual})))\in \mathbb{R}^{T},
\end{align}
where two FC layers $FC_1$ and $FC_2$ has the parameters in the dimension of $\mathbb{R}^{2T \times T}$ and $\mathbb{R}^{T \times T}$, respectively.

\subsubsection{Objective Function} 
Remarkably, it is too strict to predict the specific correct values of rPPG. Our objective is to constrain the same rhythm periodicity of the predicted rPPG and the ground-truth label. In order to meet this demand, the Pearson Correlation coefficient-based loss $\mathcal{L}_{rPPG}$ is used to compute the correlation of the predicted rPPG $s_{pre}$ with the ground-truth $s_{gt}$. The $\mathcal{L}_{rPPG}$ is formulated as:
\begin{align}
\label{eq:n_p_loss}
\mathcal{L}_{rPPG}=1-\frac{Cov(s_{pre},s_{gt})}{\sqrt{Cov(s_{pre},s_{pre})} \sqrt{Cov(s_{gt},s_{gt})}}, 
\end{align}
where $Cov(x,y)$ calcluates the covariance of two variables $x$ and $y$.

\section{Experiments} \label{experiments}
To demonstrate effectiveness of \textbf{Dual-TL}, we experiment with intra-dataset testing, cross-dataset testing, and ablation studies on four benchmark physiological measurement datasets of UBFC-rPPG~\cite{bobbia2019unsupervised}, PURE~\cite{stricker2014non}, VIPL-HR~\cite{niu2019rhythmnet}, and MMSE-HR~\cite{tulyakov2016self}. Following the previous work, the datasets of UBFC-rPPG, PURE, and VIPL-HR are used for intra-dataset testing~\cite{niu2019rhythmnet,yu2021physformer,vspetlik2018visual,lu2021dual,niu2020video,yu2020autohr}, and the MMSE-HR dataset is used for cross-dataset testing~\cite{niu2019rhythmnet,yu2021physformer,niu2020video,liu2020multi,liu2021efficientphys,yu2019remote1}.

\newcommand{\tabincell}[2]{\begin{tabular}{@{}#1@{}}#2\end{tabular}}
\begin{table*}[!t]\small
\centering
\caption{Statistics of four physiological measurement datasets (UBFC-rPPG~\cite{bobbia2019unsupervised}, PURE~\cite{stricker2014non}, VIPL-HR~\cite{niu2019rhythmnet}, and MMSE-HR~\cite{tulyakov2016self}).} 
\label{table:dataset}
\resizebox{1.0\linewidth}{!}{
\begin{tabular}{lcccccccccc}
\toprule[1pt]
\multirow{2}{*}{Dataset} & \multirow{2}{*}{Year} & \multicolumn{2}{c}{Subject} & \multirow{2}{*}{\tabincell{c}{Lighting \\Condition}} & \multirow{2}{*}{\tabincell{c}{Head \\Movement}} & \multirow{2}{*}{\tabincell{c}{Camera \\Diversity}} & \multirow{2}{*}{\tabincell{c}{Video\\\#Num.}} & \multirow{2}{*}{\tabincell{c}{Video\\Length}} & \multirow{2}{*}{\tabincell{c}{Video Frame\\Rate}} & \multirow{2}{*}{\tabincell{c}{PPG Signal \\Sampling Rate}}\\
\cline{3-4}
& & Male & Female\\
\midrule
UBFC-rPPG~\cite{bobbia2019unsupervised} & 2019 & 31 & 11 & GS & S & W & 42 & 60s & 30fps & 30Hz\\
PURE~\cite{stricker2014non} & 2014 & 8 & 2 & GS & S/SM/T & W & 60 & 60s & 30fps & 60Hz\\
MMSE-HR~\cite{tulyakov2016self} & 2016 & 17 & 23 & GS & S/E & W & 102 & 30$ \sim $60s & 25fps & 1kHz\\
VIPL-HR~\cite{niu2019rhythmnet} & 2018 & 79 & 28 & GS/DS/BS & S/LM/T & W/P & 2378 & 30s & 25$ \sim $30fps & 60Hz\\
\bottomrule[1pt]
\end{tabular}}
\begin{tablenotes}
\footnotesize
\item {$^*$ GS = General Scenario, DS = Dim Scenario, BS = Bright Scenario, S = Stable, SM = Slight Movement, LM = Large Movement, T = Talking, E = Expression, W = Web Camera, and P = Smart Phone Camera}
\end{tablenotes}
\end{table*}

\subsection{Datasets and Implementation Details}
\subsubsection{\textbf{Datasets}} 
\begin{figure}[!t]
\centering
\includegraphics[width=0.9\linewidth]{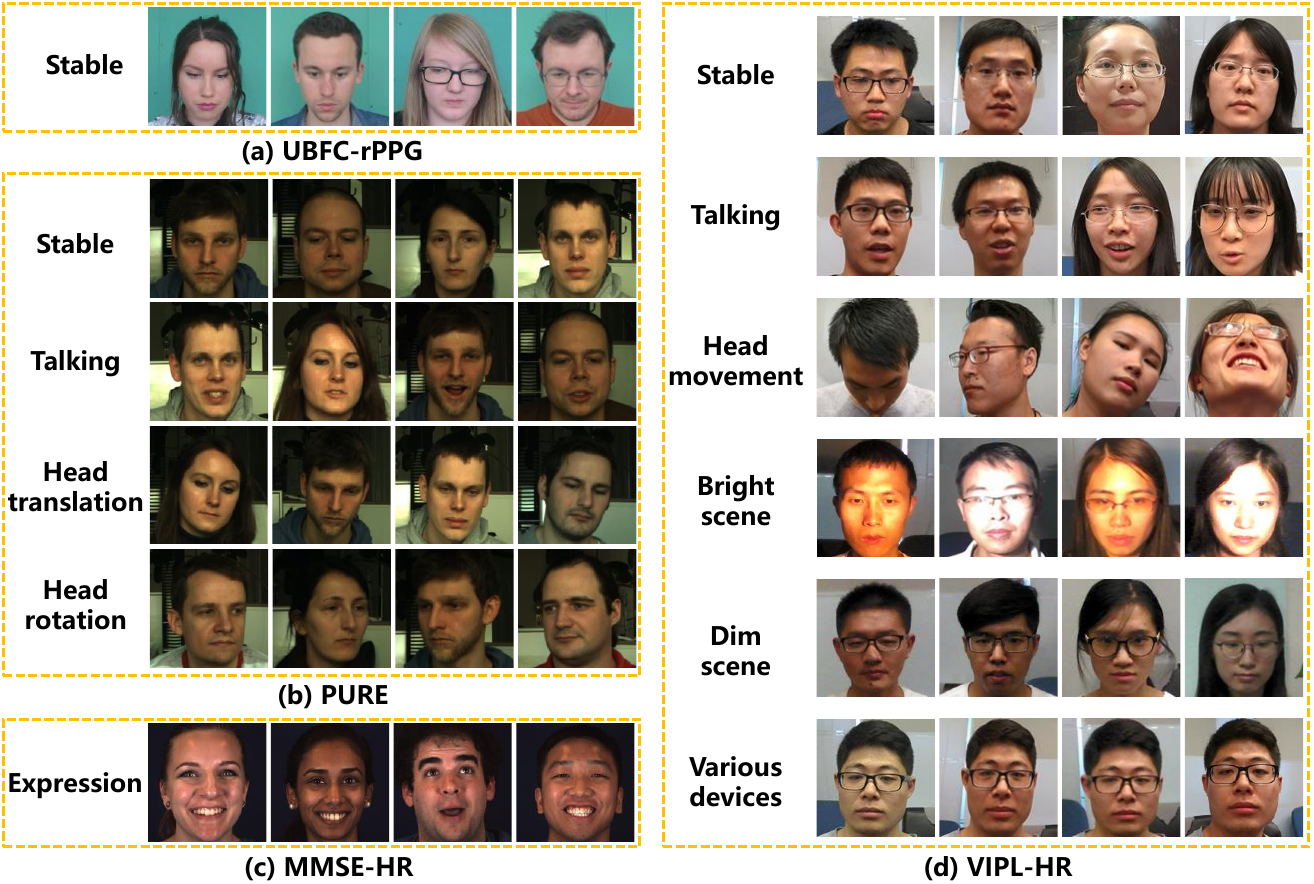}
\caption{Typical samples from different physiological measurement datasets: (a) UBFC-rPPG~\cite{bobbia2019unsupervised}, (b) PURE~\cite{stricker2014non}, (c) MMSE-HR~\cite{tulyakov2016self}, and (d) VIPL-HR~\cite{niu2019rhythmnet}. By comparison, the UBFC-rPPG is a stable dataset collected in a clean indoor scenario. While other datasets (PURE, MMSE-HR, and VIPL-HR) involve various less-constrained scenarios, such as different races, head movements, illumination changes, \etc.}
\label{fig:four_datasets}
\end{figure}

We report the statistics of four benchmark datasets for rPPG-based HR estimation in Tabel~\ref{table:dataset}. Each dataset has its own acquisition attributes of gender, lighting conditions, head movement, data size, \etc. Some representative samples from these datasets are displayed in Fig.~\ref{fig:four_datasets}. 

\textbf{\emph{UBFC-rPPG}}~\cite{bobbia2019unsupervised} consists of 42 facial videos 
performed by 42 subjects. It is a stable dataset collected in an indoor scenario. There are a training set of 30 subjects' videos and a testing set of 12 subjects' videos~\cite{lu2021dual}. 
\textbf{\emph{PURE}}~\cite{stricker2014non} is focused on head movements during physiological measurement. It consists of 60 one-minute-long videos performed by 10 subjects with six kinds of activities, including stable, talking, slow translation,
fast translation, small head rotation, and medium head rotation. The training set contains 36 videos of six subjects and the testing set contains 24 videos of the remaining four subjects~\cite{vspetlik2018visual}. 
\textbf{\emph{MMSE-HR}}~\cite{tulyakov2016self} consists of 102 facial videos. In particular, the interviewed subjects include 17 males and 23 females from diverse races. This dataset is merely used for cross-dataset testing and is entirely taken as a testing set. 
\textbf{\emph{VIPL-HR}}~\cite{niu2019rhythmnet} is a large-scale dataset in this task, including 2,378 RGB facial videos performed by 79 males and 28 females. 
The video samples are collected by four different equipment devices in nine less-constrained scenarios (\ie, stable, motion, talking, dark, bright, long distance, exercise, phone stable, and phone motion scenarios). Following the test protocol in~\cite{niu2019rhythmnet,yu2021physformer,niu2018vipl}, we use a five-fold subject-exclusive evaluation on VIPL-HR.

\subsubsection{\textbf{Implementation Details}}
Following the practice in this task~\cite{niu2019rhythmnet,niu2020video,niu2018vipl}, each video is split into multiple segments by every 0.5s, with approximately 300 frames per segment. For each subject in the video, we use OpenFace~\cite{niu2020video,baltrusaitis2018openface} to obtain 63 facial ROI combinations. The MSTmap of each segment is in the size of $N\times C\times T$. In this work, there are $N=63$, $T=300$ and $C$ = 3 or 6, where $C$ is 3, 3, and 6 under the color spaces of $\{R,G, B\}$, $\{Y,U, V\}$, and $\{R,G, B, Y,U, V\}$, respectively. Besides, a max-min normalization is performed on the MSTmap as stated in~\cite{niu2020video}, which uniformly scales the values of the MSTmap into [0, 255]. 

For model implementation, the learnable tokens ${\bf t}_{s}$ and ${\bf t}_{t}$ are initialized randomly. We set the Transformer-Encoder with six layers and four multi-heads as in ViT~\cite{dosovitskiy2020image}. 
The two fully-connected layers in the rPPG regression are set with the dimensions of $\mathbb{R}^{600 \times 600}$ and $\mathbb{R}^{600 \times 300}$, respectively.
Finally, the model outputs a $300$-$dim$ rPPG signal. We train the model via an Adam optimizer with a learning rate of 0.0001 and batch size of 128. The training epoch is set to 20 for the two datasets of UBFC-rPPG and PURE, and to 50 for the large dataset VIPL-HR, respectively. 
Here, Since the datasets used in this work are all collected for rPPG-based HR estimation, we have to convert the rPPG signal into HR values. 
Specifically, we follow previous work (\eg, EfficientPhys~\cite{liu2021efficientphys}, Dual-GAN~\cite{lu2021dual}) and apply a 1st-order Butterworth filter to the rPPG signal with a cutoff frequency range of [0.75Hz, 2.5Hz], \ie, within [45, 150] beats per minute. Then, we run the peak detection and Fast Fourier Transform (FFT) to estimate HR on each video segment. Finally, we calculate a video-level HR by averaging HRs from all video segments.

\subsubsection{\textbf{Evaluation Metrics}}
Five representative metrics 
are used for evaluation~\cite{niu2019rhythmnet,yu2021physformer,lu2021dual,niu2020video}, namely {mean absolute error} (MAE), {root mean square error} (RMSE), {mean error rate} (MER), {standard deviation of the error} (STD), and {Pearson correlation coefficient} ($r$). 
Noteworthy, all evaluations are conducted on HR values rather than rPPG signals. Taking $r$ as an example here, $r$ is used to compute the correlation between the predicted HR sequence $y_{pre}$ and the ground-truth HR sequence $y_{gt}$ of a video as bellow:
\begin{align}
r({y_{pre},y_{gt}})=\frac{Cov(y_{pre},y_{gt})}{\sigma_{y_{pre}}\cdot \sigma_{y_{gt}}} ,
\end{align}
where $\sigma$ denotes the standard deviation of a variable.
In other words, we evaluate the $r$ at the video level, where $r \in [-1,1]$. The case of $r$ = 0 indicates the irrelevance between the HR value sequences of two videos. When $r$ is 1 or -1, the predicted HR value $y_{pre}$ and the ground-truth HR value $y_{gt}$ are positively or negatively correlated.  

\subsection{Intra-dataset Testing} \label{intra-dataset testing}
\subsubsection{\textbf{Experimental Results on UBFC-rPPG}~\cite{bobbia2019unsupervised}}
Experimental results compared with state-of-the-art methods are listed in Table~\ref{table:UBFC-rppg}. UBFC-rPPG~\cite{bobbia2019unsupervised} is a pure dataset with normal illumination and few head movements. 
Even this, we can obverse that traditional methods (including GREEN~\cite{verkruysse2008remote}, ICA~\cite{poh2010non}, POS~\cite{wang2016algorithmic}, and CHROM~\cite{de2013robust}) perform significantly worse than deep learning (DL)-based methods (\eg, PulseGAN~\cite{song2021pulsegan}, Siamese-rPPG~\cite{tsou2020siamese}, and Dual-GAN~\cite{lu2021dual}). 

Our model \textbf{Dual-TL} is a DL-based method. To validate the effectiveness of our dual-path model, we also test our single path model, namely only \textbf{S-TL} or \textbf{T-TL} (introduced in Section~\ref{Dual-path TokenLearner}). 
Remarkably, our approach with only \textbf{S-TL} or \textbf{T-TL} path has already outperformed the 3D CNN-based method (\ie, Siamese-rPPG~\cite{tsou2020siamese} with high computation cost) and the GAN-based methods (\ie, PulseGAN~\cite{song2021pulsegan} and Dual-GAN~\cite{lu2021dual}).
Equipped with both \textbf{S-TL} and \textbf{T-TL}, the full model \textbf{Dual-TL} has a further performance breakthrough, which achieves the best RMSE of 0.41 bpm and has 38.8$\%$ improvement compared with the state-of-the-art method Dual-GAN~\cite{lu2021dual} (0.67 bpm).
In this stable scenario, spatial TokenLearner \textbf{S-TL} performs slightly better than temporal TokenLearner \textbf{T-TL}, such as at the RMSE of 0.50 bpm \vs 0.58 bpm. Anyway, these results demonstrate the effectiveness of our approach in spatiotemporal learning of physiological signals.

\subsubsection{\textbf{Experimental Results on PURE}~\cite{stricker2014non}} To verify the motion robustness of our model, we conduct experiments on the PURE~\cite{stricker2014non} dataset with head movement scenarios. Actually, most head movements in the PURE dataset are relatively single and slow. As shown in Table~\ref{table:PURE}, traditional methods have achieved good results, \eg, 2SR~\cite{de2014improved} and CHROM~\cite{de2013robust}. 
For the DL-based methods, HR-CNN~\cite{vspetlik2018visual} is reported as the first deep learning method that outperforms traditional methods on this dataset. 
Dual-GAN~\cite{lu2021dual} utilizes a dual generative adversarial network to differentiate the rPPG signal and noise distribution, which historically reduces the MAE value to below 1. rPPGNet~\cite{hu2021rppg} proposes a spatiotemporal attention mechanism to eliminate head movement noises and achieves MAE of 0.74 bpm and $r$ of 1.00. Notably, our approach surpasses existing methods by a large margin, establishing new state-of-the-art results (MAE of 0.37 bpm and RMSE of 0.68 bpm). Merely \textbf{S-TL} path has already outperformed all the previous methods. Moreover, on this dataset, \textbf{T-TL} path excels at distinguishing the motion variation, 
performing better than \textbf{S-TL}. These results show that our dual-path TokenLearner can effectively and adaptively explore and balance the spatial and temporal learning of facial videos, resulting in robust HR estimation performances.

\begin{table}[!t]
\centering
\caption{
Performance comparison of our proposed method and state-of-the-art methods on the UBFC-rPPG~\cite{bobbia2019unsupervised} dataset. The best results are highlighted in \textbf{bold}.} \label{table:UBFC-rppg}
\resizebox{0.9\linewidth}{!}{
\small
\begin{tabular}{lcccc}
\toprule[1pt]
Method &  MAE $\downarrow$  & RMSE $\downarrow$ & MER $\downarrow$ &
$r$ $\uparrow$\\
\midrule
\multicolumn{5}{c}{\textit{Traditional Method}} \\ \midrule
GREEN~\cite{verkruysse2008remote} & 7.05 & 14.41 & 7.82\% & 0.62\\
ICA~\cite{poh2010non} & 5.17 & 11.76 & 5.30\% & 0.65\\
POS~\cite{wang2016algorithmic}  & 4.05 & 8.75 & 4.21\% & 0.78\\
CHROM~\cite{de2013robust} & 2.37 & 4.91 & 2.46\% & 0.89\\
\midrule
\multicolumn{5}{c}{\textit{Deep Learning-based Method}} \\ \midrule
SynRhythm~\cite{niu2018synrhythm} & 5.59 & 6.82 & 5.50\% & 0.72\\
MAICA~\cite{macwan2019heart} & 3.34 & - & - & - \\
SKIN-TISSUE~\cite{bobbia2019unsupervised} & - & 2.39 & - & - \\
PulseGAN~\cite{song2021pulsegan} & 1.19 & 2.10 & 1.24\% & 0.98\\
Siamese-rPPG~\cite{tsou2020siamese} & 0.48 & 0.97 & - & -\\
Dual-GAN~\cite{lu2021dual}& 0.44 & 0.67 & 0.42\% & 0.99\\
\midrule
\textbf{S-TL(Ours)} & 0.25 & 0.50 & 0.24\% & 0.99\\
\textbf{T-TL(Ours)} & 0.33 & 0.58 & 0.36\% & 0.99\\
\textbf{Dual-TL(Ours)} & \textbf{0.17} & \textbf{0.41} & \textbf{0.17}\% & \textbf{0.99}\\
\bottomrule[1pt]
\end{tabular}}
\end{table}

\begin{table}[!t]
\renewcommand\arraystretch{0.9}
\centering
\caption{Performance comparison of our proposed method and state-of-the-art methods on the PURE~\cite{stricker2014non} dataset.}\label{table:PURE}
\resizebox{0.4\textwidth}{!}{
\begin{tabular}{lccc}
\toprule[1pt]
Method &  MAE $\downarrow$ & RMSE $\downarrow$  & 
$r$ $\uparrow$\\
\midrule
\multicolumn{4}{c}{\textit{Traditional Method}} \\ \midrule
Li2014~\cite{li2014remote} & 28.22 & 30.96 & -0.38\\
POS~\cite{wang2016algorithmic} & 3.14 & 10.57 & 0.95\\
2SR~\cite{de2014improved}  & 2.44 & 3.06 & 0.98\\
CHROM~\cite{de2013robust} & 2.07 & 2.50 & 0.99\\
\midrule
\multicolumn{4}{c}{\textit{Deep Learning-based Method}} \\ \midrule
HR-CNN~\cite{vspetlik2018visual} & 1.84 & 2.37 & 0.98\\ 
PhysNet~\cite{yu2019remote1} & 1.90 & 3.44 & 0.98\\
Dual-GAN~\cite{lu2021dual} & 0.82 & 1.31 & 0.99\\
rPPGNet~\cite{hu2021rppg} & 0.74 & 1.21 & \textbf{1.00}\\
\midrule
\textbf{S-TL(Ours)} & 0.50 & 0.92 & 0.99\\
\textbf{T-TL(Ours)} & 0.42 & 0.71 & 0.99\\
\textbf{Dual-TL(Ours)} & \textbf{0.37} & \textbf{0.68} & 0.99\\
\bottomrule[1pt]
\end{tabular}}
\end{table}

\begin{table}[!t]
\small
\centering
\caption{Performance comparison of our proposed method and state-of-the-art methods on the VIPL-HR~\cite{niu2019rhythmnet} dataset.}
\label{table:VIPL-HR}
\resizebox{0.9\linewidth}{!}{
\begin{tabular}{lcccc}
\toprule[1pt]
Method &  STD $\downarrow$ & MAE $\downarrow$  & RMSE $\downarrow$  & 
$r$ $\uparrow$\\
\midrule
\multicolumn{5}{c}{\textit{Traditional Method}} \\ \midrule
SAMC~\cite{tulyakov2016self} & 18.0 & 15.9 & 21.0 & 0.24\\
POS~\cite{wang2016algorithmic} & 15.3 & 11.5 & 17.2 & 0.24\\
CHROM~\cite{de2013robust} & 15.1 & 11.4 & 16.9 & 0.27\\
\midrule
\multicolumn{5}{c}{\textit{Deep Learning-based Method}} \\ \midrule
I3D~\cite{carreira2017quo} & 15.9 & 12.0 & 15.9 & 0.29\\
DeepPhy~\cite{chen2018deepphys} & 13.6 & 11.0 & 13.8 & 0.72\\
PhysNet~\cite{yu2019remote1} & 14.9 & 10.8 & 14.8 & 0.20\\
RhythmNet~\cite{niu2019rhythmnet} & 8.11 & 5.30 & 8.14 & 0.76\\
ST-Attention~\cite{niu2019robust} & 7.99 & 5.40 & 7.99 & 0.66\\
CVD~\cite{niu2020video} & 7.92 & 5.02 & 7.97 & 0.79\\
Dual-GAN~\cite{lu2021dual} & 7.63 & 4.93 & 7.68 & 0.81\\
PhysFormer~\cite{yu2021physformer} & 7.74 & 4.97 & 7.79 & 0.78\\
NEST~\cite{luneuron} & 7.49 & 4.76 & 7.51 & \textbf{0.84}\\
\midrule
\textbf{S-TL(Ours)} & 6.72 & 4.91 & 7.64 & 0.63\\
\textbf{T-TL(Ours)} & 6.68 & 4.51 & 7.14 & 0.66\\
\textbf{Dual-TL(Ours)} & \textbf{6.38} & \textbf{4.36} & \textbf{6.92} & 0.69\\
\bottomrule[1pt]
\end{tabular}}
\end{table}

\begin{table}[!t]
\renewcommand\arraystretch{1.0}
\centering
\caption{Experimental results of our method and the state-of-the-art methods on cross-dataset testing.}
\label{table:cross-dataset}
\resizebox{1.0\linewidth}{!}{
\begin{tabular}{lcccc}
\toprule[1pt]
{Setting} & {Method} & {MAE $\downarrow$}  & {RMSE $\downarrow$} & 
{$r$ $\uparrow$}\\
\midrule
\multirow{10}{*}{\begin{tabular}[c]{@{}c@{}}{VIPL-HR}~\cite{niu2019rhythmnet}\\ {$\rightarrow$MMSE-HR}~\cite{tulyakov2016self}\end{tabular}} &
{Li2014}~\cite{li2014remote} & {-} & {19.95} & {0.38}\\
& {CHROM}~\cite{de2013robust} & {-} & {13.97} & {0.55}\\
& {SAMC}~\cite{tulyakov2016self} & {-} & {11.37} & {0.71}\\
& {PhysNet}~\cite{yu2019remote1} & {-} & {13.25} & {0.44}\\
& {ST-Attention}~\cite{niu2019robust} & {-} & {10.10} & {0.64}\\
& {RhythmNet}~\cite{niu2019rhythmnet} & {-} & {7.33} & {0.78}\\
& {CVD}~\cite{niu2020video} & {5.02} & {7.97} & {0.79}\\
& {AutoHR}~\cite{yu2020autohr}    & {-}   & {5.87}   & {0.89}   \\
& {PhysFormerr~\cite{yu2021physformer}} & {2.84} & {5.36} & {0.92}\\ 
& {\textbf{Dual-TL (Ours)}}  & {\textbf{2.25}} & {\textbf{5.11}} & {\textbf{0.93}} \\\midrule
\multirow{4}{*}{\begin{tabular}[c]{@{}c@{}}{PURE}~\cite{stricker2014non}\\ {$\rightarrow$MMSE-HR}~\cite{tulyakov2016self}\end{tabular}} &
{CHROM}~\cite{de2013robust} & {5.72}   & {12.69}  & {0.58} \\
& {POS}~\cite{wang2016algorithmic} & {4.98}   & {13.11}  & {0.53}   \\
& {CVD}~\cite{niu2020video} & {4.08}   & {7.03}   & {0.84}   \\
& {\textbf{Dual-TL (Ours)}} & {\textbf{2.51}} &  {\textbf{5.27}} & {\textbf{0.93}} \\ \midrule 
\multirow{4}{*}{\begin{tabular}[c]{@{}c@{}}{MMSE-HR}~\cite{tulyakov2016self}\\ {$\rightarrow$PURE}~\cite{stricker2014non}\end{tabular}} &
{CHROM}~\cite{de2013robust} & {3.25}   & {12.92}  & {0.84} \\
& {POS}~\cite{wang2016algorithmic} & {2.83}   & {12.49}  & {0.85}   \\
& {CVD}~\cite{niu2020video} & {2.75}   & {3.98}   & {0.98}   \\
& {\textbf{Dual-TL (Ours)}} & {\textbf{2.18}} & {\textbf{3.23}} & {\textbf{0.98}} \\  \midrule
\multirow{4}{*}{\begin{tabular}[c]{@{}c@{}}{UBFC-rPPG}~\cite{bobbia2019unsupervised}\\ {$\rightarrow$PURE}~\cite{stricker2014non}\end{tabular}} &
{PhysNet}~\cite{yu2019remote1}  & {31.45} & {39.25}  & {-} \\
& {TS-CAN}~\cite{liu2020multi}  & {16.77}   & {31.28}  & {-}   \\
& {PhysFormerr}~\cite{yu2021physformer} & {23.63}  & {30.70}   & {-}   \\
& {\textbf{Dual-TL (Ours)}}   & {\textbf{14.12}} & {\textbf{23.47}} & {\textbf{0.23}} \\ 
\bottomrule[1pt]
\end{tabular}}
\end{table}

\begin{table*}[!t]
\centering
\caption{Ablation studies of S-TL, T-TL, and Dual-TL in the term of metric RMSE (bpm $\downarrow$) on each subcategory of VIPL-HR~\cite{niu2019rhythmnet} dataset.} 
\label{table:token}
\resizebox{0.95\linewidth}{!}{
\begin{tabular}{c|ccccccccc|c}
\toprule[1pt]
Method & Stable & Motion & Talking & Dark & Bright & Long Distance & Exercise & Phone Stable & Phone Motion & All \\ \midrule
\textbf{S-TL w/o token} & 4.95 & 6.34 & 6.77 & 4.66 & 6.10 & 4.92 & 15.39 & 5.28 & 7.80 & 7.71 \\
\textbf{S-TL} & 4.21 & 6.93 & 6.65 & 4.45 & 6.01 & 5.17 & 15.26 & 4.69 & 7.63 & 7.64 \\
\midrule
\textbf{T-TL w/o token} & 6.69 & 6.91 & 6.43 & 6.26 & 7.03 & 6.35 & 15.04 & 6.09 & 7.15 & 8.11\\
\textbf{T-TL} & 4.38 & 6.14 & 5.70 & 5.14 & 5.50 & 4.59 & 14.20 & 4.81 & 8.52 & 7.14\\
\midrule
\textbf{Dual-TL w/o token} & 4.78 & 6.54 & 6.31 & 4.93 & 5.31 & 4.70 & 14.14 & 4.32 & 7.65 & 7.21\\
\textbf{Dual-TL} & 4.08 & 6.26 & 6.27 & 4.33 & 5.13 & 4.23 & 13.66 & 4.30 & 7.25 & 6.92\\
\bottomrule[1pt]
\end{tabular}}
\end{table*}

\begin{table}[t]
\centering
\caption{Ablation study of the color space for building MSTmap on the VIPL-HR~\cite{niu2019rhythmnet} dataset.}
\label{table:color_space}
\begin{tabular}{l|cccc}
\toprule[1pt]
Color Space &  STD $\downarrow$ & MAE $\downarrow$  & RMSE $\downarrow$ & 
$r$ $\uparrow$\\
\midrule[0.3pt]
RGB & 7.94 & 5.12 & 7.72 & 0.57\\
YUV & \textbf{6.38} & \textbf{4.36} & \textbf{6.92} & \textbf{0.69}\\
RGB+YUV & 7.38 & 4.87 & 7.57 & 0.59\\
\bottomrule[1pt]
\end{tabular}
\end{table}

\begin{table}[t]
\caption{Ablation studies of ReLU and GELU activation functions in the rPPG regression head on the VIPL-HR dataset.}
\centering
\begin{tabular}{c|cccc}
\hline
{Method} & {STD$\downarrow$} & {MAE$\downarrow$}  & {RMSE$\downarrow$} & {r$\uparrow$}    \\ \hline
{Dual-TL (ReLU)}   & {6.46}  & {4.38} & {7.01} & {0.69} \\
{Dual-TL (GELU)}   & {\textbf{6.38}}  & {\textbf{4.36}} & {\textbf{6.92}} & {\textbf{0.69}} \\ \hline
\end{tabular}
\label{tab:relu}
\end{table}

\subsubsection{\textbf{Experimental results on VIPL-HR}~\cite{niu2019rhythmnet}} 
Table~\ref{table:VIPL-HR} lists the state-of-the-art performances on the VIPL-HR datatset~\cite{niu2019rhythmnet}. Our method achieves the lowest metrics, \eg, STD (6.38 bpm $\downarrow$), MAE (4.36 bpm $\downarrow$), and RMSE (6.92 bpm $\downarrow$). 
Compared to Dual-GAN~\cite{lu2021dual} and NEST~\cite{luneuron}, our performance improvement is significantly obvious with the drop of MAE by 11.6$\%$ and 8.4$\%$, respectively.

The VIPL-HR dataset is a challenging large-scale dataset, which contains nine less-constrained noisy scenarios, such as head movements, talking, illumination variations, and acquisition devices, \etc. As shown in Table~\ref{table:VIPL-HR}, traditional hand-crafted methods (\eg, SAMC~\cite{tulyakov2016self}, POS~\cite{wang2016algorithmic}, and CHROM~\cite{de2013robust}) perform poorly in these noisy scenarios. Deep neural networks with strong modeling capabilities driven by sufficient (large-scale) data promote performance improvements obviously, such as DeepPhy~\cite{chen2018deepphys}, PhysNet~\cite{yu2019remote1}, and RhythmNet~\cite{niu2019rhythmnet}.

\begin{table}[t]
\small
\centering
\caption{Ablation studies of the layer and multi-head of Transformer-Encoder 
on the  VIPL-HR~\cite{niu2019rhythmnet} dataset.}
\label{table:head_layer}
\resizebox{0.9\linewidth}{!}{
\begin{tabular}{l|l|cccc}
\toprule[1pt]
Hyperparameters & Setting &  STD $\downarrow$ & MAE $\downarrow$ & RMSE $\downarrow$ & 
$r$ $\uparrow$\\
\midrule
\multirow{5}*{$head=4$}&$layer$ = 2 & 8.78 & 5.94 & 8.69 & 0.45\\
~&$layer$ = 4 & 6.97 & 4.82 & 7.37 & 0.23\\
~&\textbf{$layer$ = 6} & \textbf{6.38} & \textbf{4.36} & \textbf{6.92} & \textbf{0.69}\\
~&$layer$ = 8 & 6.58 & 4.50 & 7.10 & 0.67\\
~&$layer$ = 10 & 6.91 & 4.67 & 7.31 & 0.64\\
\midrule
\multirow{6}*{$layer=6$}&$head$ = 1 & 6.67 & 4.51 & 7.42 & 0.63\\
~&$head$ = 2 & 6.59 & 4.47 & 7.34 & 0.64\\
~&$head$ = 3 & 6.43 & 4.41 & 7.11 & 0.68\\
~&\textbf{$head$ = 4} & \textbf{6.38} & \textbf{4.36} & \textbf{6.92} & \textbf{0.69}\\
~&$head$ = 5 & 6.42 & 4.39 & 7.09 & 0.68\\
~&$head$ = 6 & 6.53 & 4.43 & 7.26 & 0.65\\
\bottomrule[1pt]
\end{tabular}
}
\end{table}

For the discussion of DL-based approaches, we analyze that CNN-based models such as RhythmNet~\cite{niu2019rhythmnet}, CVD~\cite{niu2020video}, Dual-GAN~\cite{lu2021dual}, NEST~\cite{luneuron} are constrained by the local receptive field of convolutional operation. The only Transformer-based work test on this dataset -- PhysFormer~\cite{yu2021physformer} --
explores video representation. In contrast to these methods, our method utilizes two learnable tokens (Spatial and Temporal TokenLearner) to recover the intrinsic periodicity of rPPG. 
As excepted, our Dual-path TokenLearner achieves much better evaluation results than PhysFormer (\eg, STD of 6.75 bpm \vs 7.74 bpm, MAE of 4.64 bpm \vs 4.97 bpm, and RMSE of 7.08 bpm \vs 7.79 bpm). 
The results further verify the robustness of Dual-path TokenLearner in capturing rPPG signals in complicated environments.

\begin{figure}[t]
\centering
\includegraphics[width=1.0\linewidth]{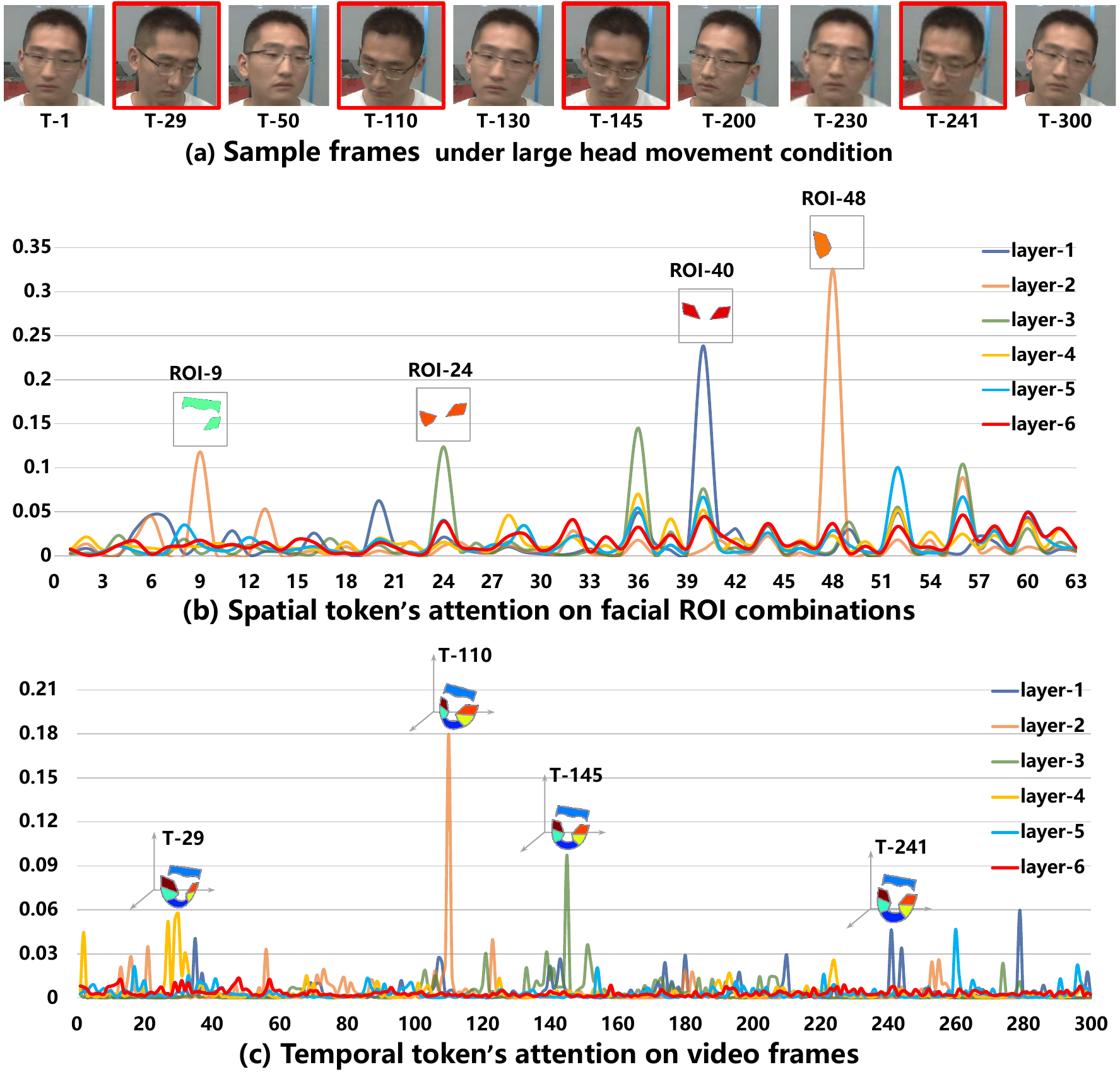}
\caption{
Visualization of 
Temporal TokenLearner's and Spatial TokenLearner's (\textbf{T-TL}'s \& \textbf{S-TL}'s) attention distribution at each Transformer-Encoder layer. (a) Some specific frames at different timestamps.
(b) Spatial token’s interactive attention with each facial ROI combination. (c) Temporal token’s interactive attention with each video frame. Note that ROI-$i$, $i \in [1,63]$, represents the $i$-th ROI combination, and T-$j$, $j \in [1,300]$, represents the $j$-th frame in the video.}
\label{fig:temporal_attention}
\end{figure}

\subsection{Cross-dataset Testing} 
The cross-dataset experiment is used to evaluate the generalizability of the model. 
Following the previous works~\cite{yu2021physformer,suncontrast}, we conduct four cross-dataset testins. As shown in Table~\ref{table:cross-dataset}, our Dual-TL achieves state-of-the-art performance across all the cross-dataset settings. For example, when compared to our method with the latest Transformer-based PhysFormer~\cite{yu2021physformer} on the cross-dataset setting of VIPL-HR$\rightarrow$MMSE-HR (taking VIPL-HR as training data and MMSE-HR as testing data), the MAE of our model is significantly reduced by 21$\%$ (from 2.84 bpm to 2.25 bpm). These results show a strong generalization ability of our method in the unknown scenario. 

\subsection{Ablation Studies}
\subsubsection{Color Space of MSTmap} \label{color_space}
A suitable color space benefits the rPPG measurement~\cite{yang2016motion}. Here, we test the color spaces of YUV, RGB, and RGB+YUV. 
As shown in Table~\ref{table:color_space}, the YUV achieves the best result with the RMSE of 7.08 bpm. 
It indicates that the YUV color space is more favorable for capturing the heart rhythm clues in our experiments. This conclusion is the same as in ~\cite{niu2019rhythmnet}. 
Thus, we apply YUV as the color space setup for building MSTmap in our experiments.

\subsubsection{Transformer-Encoder} 
We test the effect of the encoder layer and multi-head in our Transformer-Encoder. From Table~\ref{table:head_layer}, the model achieves the best performance when the layer and head numbers are set to 6 and 4, respectively. There is an interesting observation that under the setting of ``6-layer \& any head number" or ``4-head \& any layer number", our Dual-path TokenLearner is competitive with the state-of-the-art methods.
We take the 6-layer and 4-head as the optional setup. 

\subsubsection{Dual-path TokenLearner} \label{Dual-path TokenLearner}
To validate the effectiveness of \textbf{Dual-TL}, we compare it with its single path, \ie, merely \textbf{S-TL} or \textbf{S-TL}. In the setting of \textbf{S-TL}/\textbf{T-TL}, only ${\bf\hat t}_{s}$ or ${\bf\hat t}_{t}$ goes through $\bf{FFN_{reg}}$ for predicting rPPG signals. Experimental analyses on the \textbf{UBFC-rPPG} and \textbf{PURE} datasets are provided in Section~\ref{intra-dataset testing}. Here, we discuss the \textbf{VIPL-HR}~\cite{niu2019rhythmnet} dataset. As shown in Table~\ref{table:VIPL-HR}, merely \textbf{S-TL} is still superior to the advanced method Dual-GAN~\cite{lu2021dual}. \textbf{Dual-TL} further improves the performance.

Furthermore, we present the results in nine scenarios of the VIPL-HR dataset to validate the effects of the proposed \textbf{S-TL}, \textbf{T-TL}, and \textbf{Dual-TL}, respectively. 
As shown in Table~\ref{table:token}, it is clear that \textbf{S-TL} surpasses \textbf{T-TL} path by a remarkable margin in the stable scenarios (\eg, ``Stable", ``Dark", and ``Phone Stable"). However, the performance of \textbf{S-TL} is obviously worse than \textbf{T-TS} in the scenarios with movements (\eg, ``Motion" and ``Talking"). 
Besides, bright light can severely interfere with the acquisition of skin color variation for rPPG measurements; in this case, \textbf{S-TL} performs poorly. For example, \textbf{T-TL} achieves an RMSE of 5.50 bpm, which is 8.5\% better than \textbf{S-TL} in the ``Bright" scenario.
In particular, we find that the \textbf{S-TL}, \textbf{T-TL}, and \textbf{Dual-TL} consistently perform not well in the ``Exercise” scenario, such as 15.26 bpm of \textbf{S-TL}, 14.20 bpm of \textbf{T-TL}, and 13.66 bpm of \textbf{Dual-TL} in the term of RMSE. This is in line with the objective fact that ``Exercise” is the most challenging scenario.

\subsubsection{Learnable Token}
To test the effects of the two learnable tokens, we conduct ablation studies with three variants without token learning. \textbf{S-TL w/o token} means that we remove the spatial token ${\bf t_s}$ and input AvgPool (${\bf\hat X}_s^{'}$) to replace ${\bf t_s}$ in ${\bf t}_{dual}$ (refer to Eq.~\ref{eq6}) for rPPG regression, where ${\bf\hat X}_s^{'}$ denotes the new spatial video features obtained by Eq.~\ref{eq3}. 
Similar to \textbf{S-TL w/o token}, \textbf{T-TL w/o token} inputs AvgPool (${\bf\hat X}_t^{'}$) to replace ${\bf t_t}$ in ${\bf t}_{dual}$ (refer to Eq.~\ref{eq6}). As for \textbf{Dual-TL w/o token}, both tokens ${\bf t_s}$ and ${\bf t_t}$ are removed, and corresponding spatial-temporal features [${\bf\hat X}_s^{'}$, ${\bf\hat X}_t^{'}$] are input into the rPPG regression head.
As shown in Table~\ref{table:token}, the performances of these variants drop significantly.
Specifically, the RMSE of \textbf{S-TL w/o token} is larger than \textbf{S-TL} (\ie, 7.71 bpm \vs 7.64 bpm, for the RMSE, the smaller the better). Compared to \textbf{T-TL}, \textbf{T-TL w/o token} performs particularly poorly, with an 11.9\% lift in RMSE (\ie, 8.11 bpm \vs 7.14 bpm). 
We inspect the token effects in the nine noise scenarios of the VIPL-HR dataset. For dual-path learning, the tokens help \textbf{Dual-TL} achieve a large improvement by 14.6\%, 4.3\%, 0.6\%, 12.2\%, 3.4\%, 10.0\%, 3.4\%, 0.5\% and 6.5\% in the scenarios of ``Stable", ``Motion", ``Talking", ``Dark", ``Bright", ``Long Distance", ``Exercise", ``Phone Stable" and ``Phone Motion", respectively. These results demonstrate that the usage of learnable tokens is a preferable way to aggregate spatiotemporal context for boosting the robustness rPPG measurement. 

\subsubsection{GELU Activation} 
In the \emph{rPPG Regression Head}, we adopt a GELU activation layer rather than ReLU to predict the rPPG signals. We test both the GELU and ReLU in our method. As shown in Table~\ref{tab:relu}, GELU exhibits a more stable improvement than ReLU, which demonstrates its efficacy in the continuous rPPG prediction. The GELU addresses the limitation of ReLU's derivative discontinuity and can provide satisfactory performance in handling the nonlinear change of continuous data for this task.

\begin{figure*}
\centering
\includegraphics[width=0.95\linewidth]{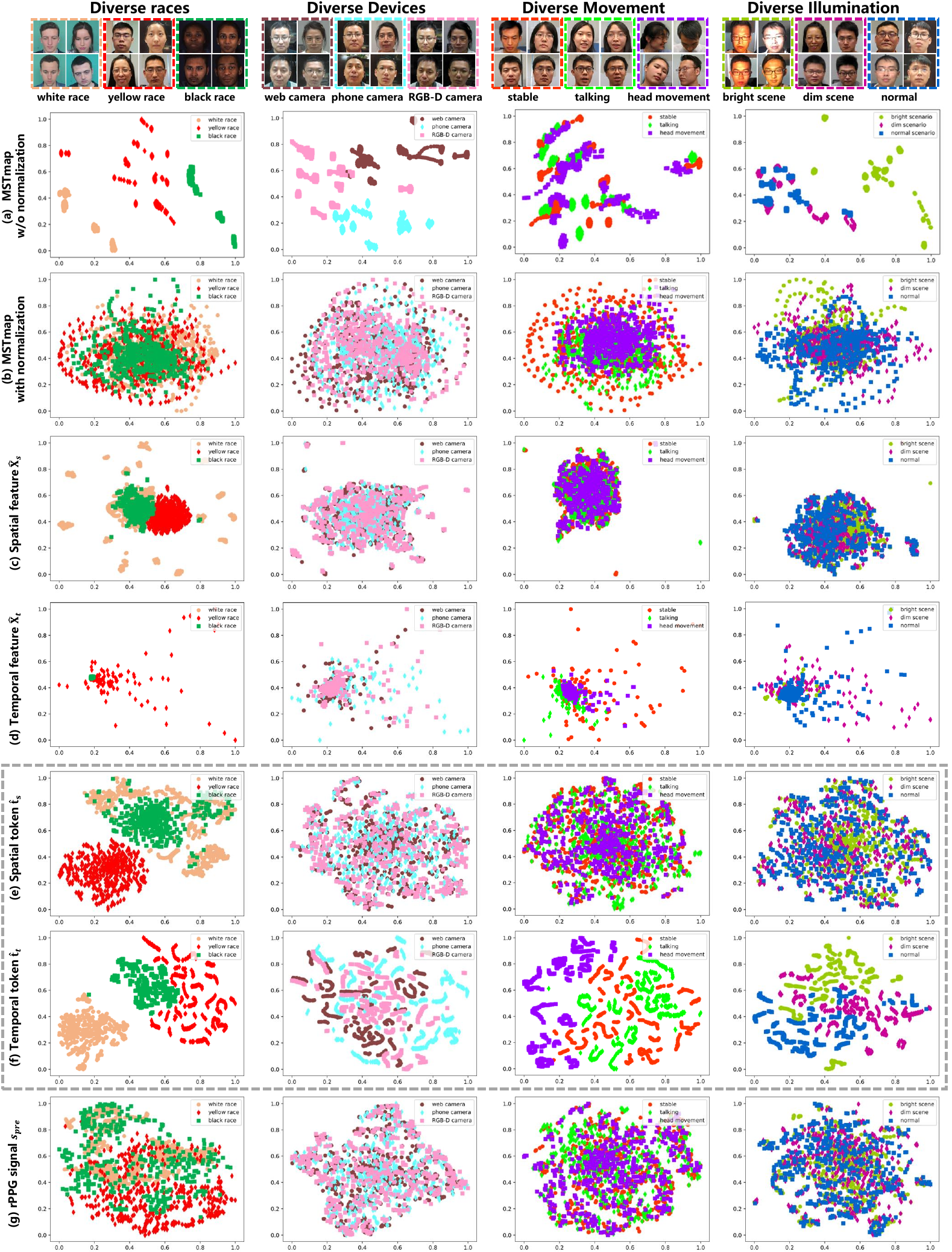}
\caption{The t-SNE~\cite{van2008visualizing} visualization of feature representations learned by Dual-path TokenLearner. The samples of noisy scenarios (``Diverse Devices", ``Diverse Movements'', and ``Diverse Illumination'') are acquired from three test subsets of VIPL-HR~\cite{niu2019rhythmnet} dataset. Due to the lack of a multi-race dataset, we merge the test sets of the  UBFC-rPPG~\cite{bobbia2019unsupervised} (white race), MMSE-HR~\cite{tulyakov2016self} (black race), and VIPL-HR~\cite{niu2019rhythmnet} (yellow race) datasets to construct the ``Diverse Races'' scenario. The video samples in the ``Diverse Races'' scenario are independently trained with their own original datasets. We display them here to show that even with different training schemes, the finally predicted rPPG signals are evenly distributed. Different colors correspond to different sub-categories.}
\label{fig:tsne}
\end{figure*}

\begin{figure*}
\centering
\includegraphics[width=1\linewidth]{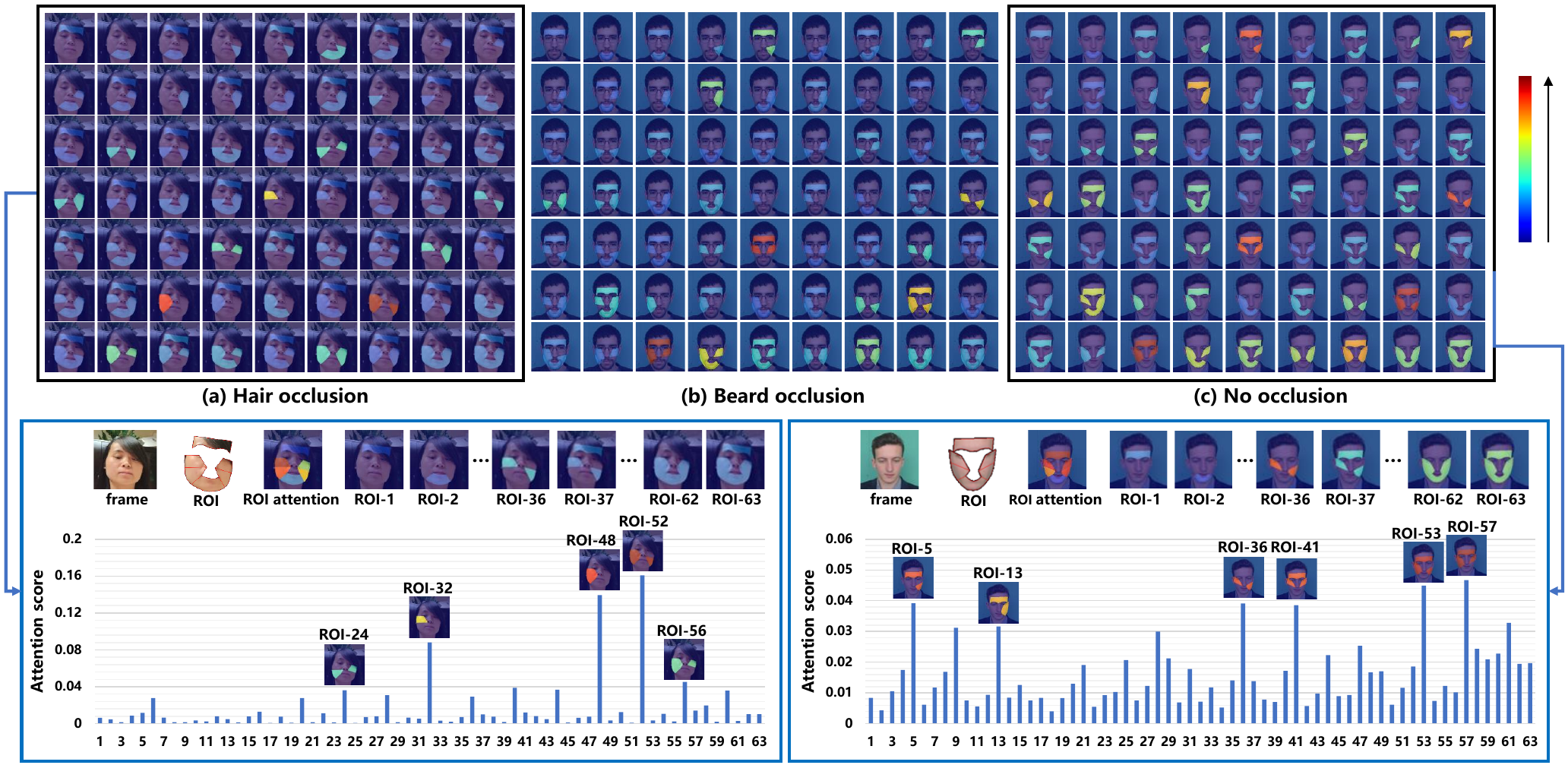}
\caption{Visualization of Spatial TokenLearner (\textbf{S-TL}) -- spatial token's average attention from all the layers of Transformer-Encoder over 63 ROI combinations. ROI-$i$ ($i$=1, 2, 3, $\cdots$, 63) represents the $i$-th ROI combination patch. Here, we display the quantitative results of three typical cases containing Hair occlusion in (a), Beard occlusion in (b), and No occlusion in (c). In each case, we use different color shades to represent the attention scores of 63 ROI combinations. }
\label{fig:spatial_attention}
\end{figure*}

\begin{figure*}
\centering
\includegraphics[width=1.0\linewidth]{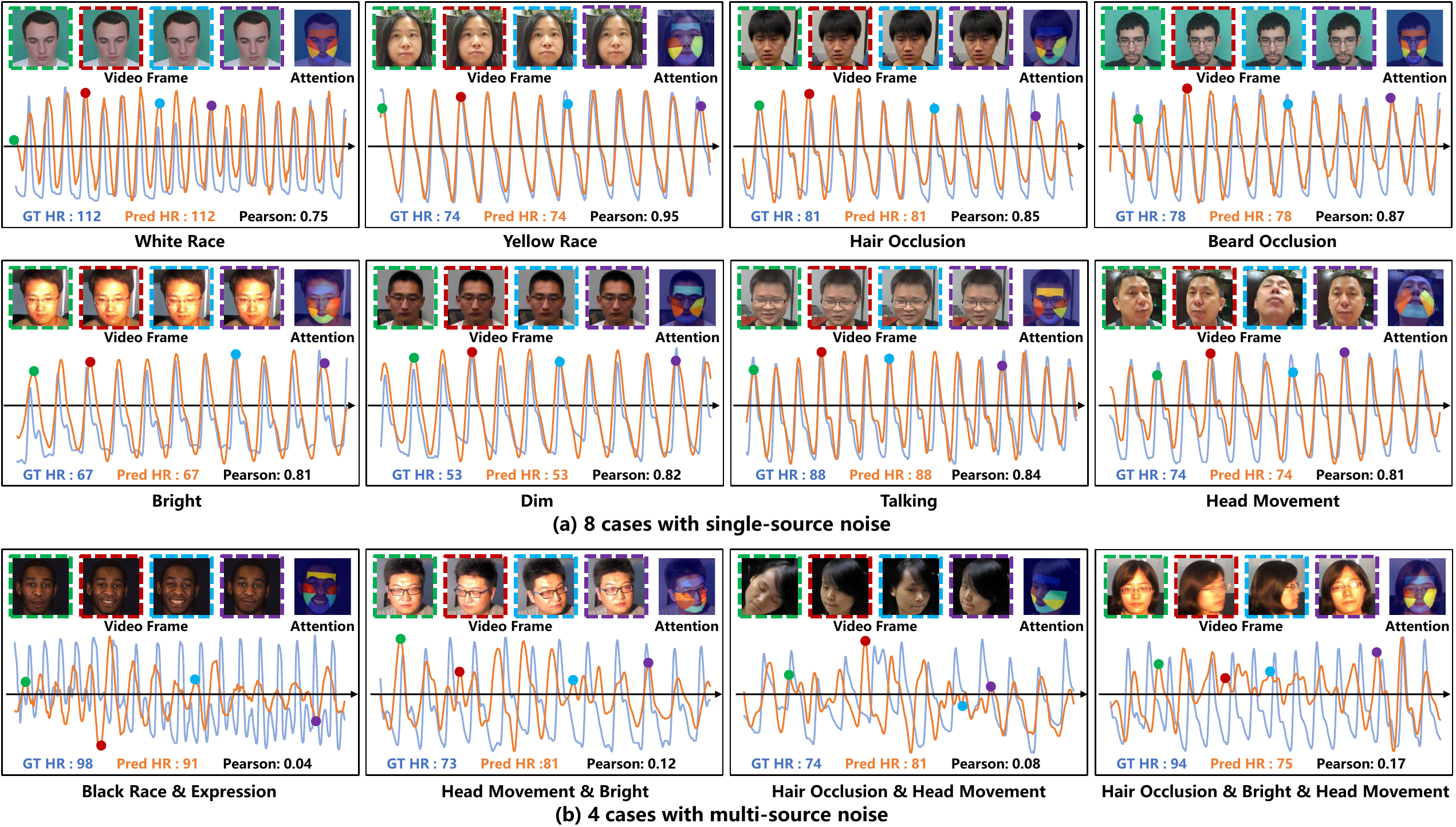}
\caption{Visualization of qualitative results in twelve challenging scenarios. Common physiological measurement noises can be summarized into single-source noises in (a) and multi-source noises in (b) according to the subcategories of four benchmark physiological datasets (UBFC-rPPG~\cite{bobbia2019unsupervised}, PURE~\cite{stricker2014non}, MMSE-HR~\cite{tulyakov2016self}, and VIPL-HR~\cite{niu2019rhythmnet}). We show the predicted rPPG signals (Pred) and corresponding ground-truth PPG signals (GT) in blue and orange curves, respectively. Some representative video frames are marked in the waveform with dots and placed on the top of the signal diagram by different color boxes. We also display the Pred and the GT HR values as well as the ``Pearson" correlation coefficient between the Pred and GT signals.}
\label{fig:qualitative results}
\end{figure*}

\subsection{Quantitative Results 
}
\subsubsection{Feature Embedding} \label{t_sne} 
To demonstrate the effectiveness of Dual-path TokenLearner, we visualize the feature distribution by using t-SNE~\cite{van2008visualizing}. As shown in Fig.~\ref{fig:tsne}, we display seven kinds of features, \ie, original MSTmap, MSTmap with Max-Min Normalization, spatial feature (${\bf\hat X}_s$), spatial token (${\bf\hat t}_s$), temporal feature (${\bf\hat X}_t$), temporal token (${\bf\hat t}_t$), and the predicted rPPG signal ($s_{pre}$). 
Regarding the noise disturbances, we choose four scenarios with diverse races (\ie, white race, yellow race, and black race), acquisition devices (\ie, web camera, phone camera, and RGB-D camera), illumination conditions (\ie, bright, dark, and normal), and movements (\ie, stable, talking, and head movement). 

From the preliminary MSTmap features in Fig.~\ref{fig:tsne} (a), the feature distribution of each subcategory is obviously scattered and influenced by individual characteristics or respective environmental factors.
After MSTmap normalization in Fig.~\ref{fig:tsne} (b), the MSTmap features are blended together. This can eliminate the color difference between different samples and control the difference into a reasonable range. 
After Transformer-based encoding, the spatial and temporal features respectively are aggregated, as shown in Figs.~\ref{fig:tsne} (c) and (e). For learnable tokens, the spatial token is relatively distributed evenly in Fig.~\ref{fig:tsne} (d), while the temporal token exhibits a strong sequentiality in Fig.~\ref{fig:tsne} (f).  We can see that the noise bias with the expectation of even distribution has been mainly addressed by spatial tokens (illumination, movements, \etc), while the effect of the temporal token is to handle the continuity and smoothness of the predicted rPPG curve. The predicted rPPG curves of video instances can be observed in Fig.~\ref{fig:qualitative results}. Indeed, the rPPG curve generated by our method is smooth.
Overall, with both effective spatial and temporal token learning, the final rPPG signals have the most uniform distribution as shown in Fig.~\ref{fig:tsne} (h). As expected, the model exhibits the ability of deindividuation and denoising in the rPPG measurement.

\subsubsection{Spatial TokenLearner (\textbf{S-TL})}
To further demonstrate the effectiveness of \textbf{S-TL}, we show the interactive weights between the spatial token and different facial ROI combinations from each Transformer-Encoder layer. Our model aims to seek suitable ROI combinations for rPPG predictions. 
From Figs.~\ref{fig:temporal_attention} (a) and (b), it attends at ROI-9, ROI-24, ROI-40, and ROI-48 with higher weight values at early layers.
In these cases, the detected ROI combinations cover the facial regions with more salient appearances such as the forehead and cheek. After several Transformer-Encoder layers, the attention weight related to each ROI combination is relatively uniform. This indicates that our model absorbs the positive effects of attentive facial ROI appearances and then covers and balances the comprehensive contribution of all the facial ROI appearances.

Furthermore, we average the attention weights of all the Transformer-Encoder layers to display individual differences. 
From this viewpoint with three examples (a) $\sim$ (c) in Fig.~\ref{fig:spatial_attention}, the model obviously prefers the ROI combinations without occlusion. As the woman with hair occlusion shown in Fig.~\ref{fig:spatial_attention} (a), the attentive weights sparsely appear and mostly happen on the cheek-related ROI combinations. 
In Fig.~\ref{fig:spatial_attention} (c), the man without occlusion has close attention weights over almost facial ROI combinations. By further observation, in this case, relatively larger attention weights are concentrated on the forehead and cheeks. It again indicates that the forehead and cheeks are more effective than the chin in reflecting the heartbeat's rhythm.
Remarkably, these quantitative results prove that our model can reasonably attend favorable face regions for rPPG-based HR estimation.

\subsubsection{Temporal TokenLearner (\textbf{T-TL})}
Here, we visualize the interactive attention of temporal tokens along the timeline from each Transformer-Encoder layer. 
As the attention curve shown in Fig.~\ref{fig:temporal_attention} (c), the sharp peak at early layers often happens at the frames with obvious and quick head movements, such as 29-th, 110-th, 145-th, and 241-th frames in Fig.~\ref{fig:temporal_attention} (a). 
As the same conclusion to the spatial token, we discover that the attention weight of each frame after the last encoder layer is relatively uniform. After implementing $L=6$ encoder layers, our model progressively avoids noisy interference.

\subsubsection{Robust HR Estimation Samples}
Finally, we discuss the noisy scenarios. We select eight single-source scenarios and four multi-source noise scenarios.
For single-source noise in Fig.~\ref{fig:qualitative results} (a), our approach predicts both accurate rPPG signals and HR values. Even though the predicted rPPG signals sometimes are not coincident with the ground truth in the scenarios such as ``White Race" and ``Bright", the periodic patterns of peaks and troughs are completely consistent. 
In other words, they have consistent fluctuation trends to express the heartbeat frequency, thereby achieving accurate HR values.
Remarkably, the multi-source noise scenario is a potential challenge for our model. 
As shown in Fig.~\ref{fig:qualitative results} (b), while multiple types of noises happen (\eg, in the scenarios of ``Head Movement \& Bright" and ``Hair Occlusion \& Bright \& Head Movement"), the curve of the predicted rPPG signal does not fit the curve of ground-truth well, but the crests between the predicted rPPG signal and ground-truth PPG signal are well aligned. This indicates the strong robustness of our method. 

\section{Conclusions}\label{conclusions}
This paper proposes a new Transformer-based framework named Dual-path TokenLearner (Dual-TL) using facial videos in a non-contact manner for remote physiological measurement. It consists of two learnable tokens, Spatial TokenLeaner (S-TL) and Temporal TokenLeaner (T-TL), that retain informative spatial and temporal contexts, respectively. The S-TL and T-TL are performed in parallel to exploit complementary spatiotemporal dependencies in facial videos and eliminate external disturbances. Experiments conducted on four physiological benchmarks demonstrate a large margin of improvement in the performance of our Dual-TL compared to previous state-of-the-arts. 
Extensive ablation studies and visualization results show the effectiveness of Dual-TL in tackling complicated noise scenarios such as illumination variations, facial occlusions, and head movements. By analysis, S-TL concentrates on the facial parts such as the forehead and cheeks that better reflect the rhythm of the heartbeat, and T-TL prefers frames with clear, unobstructed facial ROI regions. Besides, although single-source noise scenarios can be addressed well, multi-source noise scenarios are still challenges to be solved in this field.

\bibliographystyle{IEEEtran}
\bibliography{IEEEabrv}

\begin{thebibliography}{10}
\providecommand{\url}[1]{#1}
\csname url@samestyle\endcsname
\providecommand{\newblock}{\relax}
\providecommand{\bibinfo}[2]{#2}
\providecommand{\BIBentrySTDinterwordspacing}{\spaceskip=0pt\relax}
\providecommand{\BIBentryALTinterwordstretchfactor}{4}
\providecommand{\BIBentryALTinterwordspacing}{\spaceskip=\fontdimen2\font plus
\BIBentryALTinterwordstretchfactor\fontdimen3\font minus
  \fontdimen4\font\relax}
\providecommand{\BIBforeignlanguage}[2]{{%
\expandafter\ifx\csname l@#1\endcsname\relax
\typeout{** WARNING: IEEEtran.bst: No hyphenation pattern has been}%
\typeout{** loaded for the language `#1'. Using the pattern for}%
\typeout{** the default language instead.}%
\else
\language=\csname l@#1\endcsname
\fi
#2}}
\providecommand{\BIBdecl}{\relax}
\BIBdecl

\bibitem{schiweck2019heart}
C.~Schiweck, D.~Piette, D.~Berckmans, S.~Claes, and E.~Vrieze, ``Heart rate and
  high frequency heart rate variability during stress as biomarker for clinical
  depression. a systematic review,'' \emph{Psychological Medicine}, vol.~49,
  no.~2, pp. 200--211, 2019.

\bibitem{sunprivacy}
Z.~Sun and X.~Li, ``Privacy-phys: Facial video-based physiological modification
  for privacy protection,'' \emph{IEEE Signal Processing Letters}, vol.~29, pp.
  1507--1511, 2022.

\bibitem{gilgen2019rr}
R.~Gilgen-Ammann, T.~Schweizer, and T.~Wyss, ``Rr interval signal quality of a
  heart rate monitor and an ecg holter at rest and during exercise,''
  \emph{European Journal of Applied Physiology}, vol. 119, no.~7, pp.
  1525--1532, 2019.

\bibitem{pereira2020photoplethysmography}
T.~Pereira, N.~Tran, K.~Gadhoumi, M.~M. Pelter, D.~H. Do, R.~J. Lee,
  R.~Colorado, K.~Meisel, and X.~Hu, ``Photoplethysmography based atrial
  fibrillation detection: a review,'' \emph{NPJ Digital Medicine}, vol.~3,
  no.~1, pp. 1--12, 2020.

\bibitem{de2013robust}
G.~De~Haan and V.~Jeanne, ``Robust pulse rate from chrominance-based rppg,''
  \emph{IEEE Transactions on Biomedical Engineering}, vol.~60, no.~10, pp.
  2878--2886, 2013.

\bibitem{tulyakov2016self}
S.~Tulyakov, X.~Alameda-Pineda, E.~Ricci, L.~Yin, J.~F. Cohn, and N.~Sebe,
  ``Self-adaptive matrix completion for heart rate estimation from face videos
  under realistic conditions,'' in \emph{Proc. CVPR}, 2016, pp. 2396--2404.

\bibitem{niu2019rhythmnet}
X.~Niu, S.~Shan, H.~Han, and X.~Chen, ``Rhythmnet: End-to-end heart rate
  estimation from face via spatial-temporal representation,'' \emph{IEEE
  Transactions on Image Processing}, vol.~29, pp. 2409--2423, 2019.

\bibitem{kurihara2021non}
K.~Kurihara, D.~Sugimura, and T.~Hamamoto, ``Non-contact heart rate estimation
  via adaptive rgb/nir signal fusion,'' \emph{IEEE Transactions on Image
  Processing}, vol.~30, pp. 6528--6543, 2021.

\bibitem{li2018obf}
X.~Li, I.~Alikhani, J.~Shi, T.~Seppanen, J.~Junttila, K.~Majamaa-Voltti,
  M.~Tulppo, and G.~Zhao, ``The obf database: A large face video database for
  remote physiological signal measurement and atrial fibrillation detection,''
  in \emph{Proc. FG}, 2018, pp. 242--249.

\bibitem{huang2020heart}
P.-W. Huang, B.-J. Wu, and B.-F. Wu, ``A heart rate monitoring framework for
  real-world drivers using remote photoplethysmography,'' \emph{IEEE Journal of
  Biomedical and Health Informatics}, vol.~25, no.~5, pp. 1397--1408, 2020.

\bibitem{van2020camera}
M.~van Gastel, S.~Stuijk, S.~Overeem, J.~P. van Dijk, M.~M. van Gilst, and
  G.~de~Haan, ``Camera-based vital signs monitoring during sleep--a proof of
  concept study,'' \emph{IEEE Journal of Biomedical and Health Informatics},
  vol.~25, no.~5, pp. 1409--1418, 2020.

\bibitem{wang2022domain}
Z.~Wang, Z.~Wang, Z.~Yu, W.~Deng, J.~Li, T.~Gao, and Z.~Wang, ``Domain
  generalization via shuffled style assembly for face anti-spoofing,'' in
  \emph{Proc. CVPR}, 2022, pp. 4123--4133.

\bibitem{yu2021deep}
Z.~Yu, Y.~Qin, X.~Li, C.~Zhao, Z.~Lei, and G.~Zhao, ``Deep learning for face
  anti-spoofing: A survey,'' \emph{arXiv:2106.14948}, 2021.

\bibitem{verkruysse2008remote}
W.~Verkruysse, L.~O. Svaasand, and J.~S. Nelson, ``Remote plethysmographic
  imaging using ambient light.'' \emph{Optics Express}, vol.~16, no.~26, pp.
  21\,434--21\,445, 2008.

\bibitem{li2014remote}
X.~Li, J.~Chen, G.~Zhao, and M.~Pietikainen, ``Remote heart rate measurement
  from face videos under realistic situations,'' in \emph{Proc. CVPR}, 2014,
  pp. 4264--4271.

\bibitem{liu2021camera}
X.~Liu, S.~Patel, and D.~McDuff, ``Camera-based physiological sensing:
  Challenges and future directions,'' \emph{arXiv:2110.13362}, 2021.

\bibitem{yu2021facial}
Z.~Yu, X.~Li, and G.~Zhao, ``Facial-video-based physiological signal
  measurement: Recent advances and affective applications,'' \emph{IEEE Signal
  Processing Magazine}, vol.~38, no.~6, pp. 50--58, 2021.

\bibitem{yu2021physformer}
Z.~Yu, Y.~Shen, J.~Shi, H.~Zhao, P.~Torr, and G.~Zhao, ``Physformer: Facial
  video-based physiological measurement with temporal difference transformer,''
  in \emph{Proc. CVPR}, 2022, pp. 4186--4196.

\bibitem{mcduff2020advancing}
D.~McDuff, J.~Hernandez, E.~Wood, X.~Liu, and T.~Baltrusaitis, ``Advancing
  non-contact vital sign measurement using synthetic avatars,''
  \emph{arXiv:2010.12949}, 2020.

\bibitem{lewandowska2011measuring}
M.~Lewandowska, J.~Rumi{\'n}ski, T.~Kocejko, and J.~Nowak, ``Measuring pulse
  rate with a webcam—a non-contact method for evaluating cardiac activity,''
  in \emph{Proc. FedCSIS}, 2011, pp. 405--410.

\bibitem{lam2015robust}
A.~Lam and Y.~Kuno, ``Robust heart rate measurement from video using select
  random patches,'' in \emph{Proc. ICCV}, 2015, pp. 3640--3648.

\bibitem{poh2010advancements}
M.-Z. Poh, D.~J. McDuff, and R.~W. Picard, ``Advancements in noncontact,
  multiparameter physiological measurements using a webcam,'' \emph{IEEE
  Transactions on Biomedical Engineering}, vol.~58, no.~1, pp. 7--11, 2010.

\bibitem{wang2016algorithmic}
W.~Wang, A.~C. Den~Brinker, S.~Stuijk, and G.~De~Haan, ``Algorithmic principles
  of remote ppg,'' \emph{IEEE Transactions on Biomedical Engineering}, vol.~64,
  pp. 1479--1491, 2016.

\bibitem{poh2010non}
M.-Z. Poh, D.~J. McDuff, and R.~W. Picard, ``Non-contact, automated cardiac
  pulse measurements using video imaging and blind source separation.''
  \emph{Optics Express}, vol.~18, no.~10, pp. 10\,762--10\,774, 2010.

\bibitem{liproposal}
K.~Li, D.~Guo, and M.~Wang, ``Proposal-free video grounding with contextual
  pyramid network,'' in \emph{Proc. AAAI}, vol.~35, no.~3, 2021, pp.
  1902--1910.

\bibitem{guodadnet}
D.~Guo, K.~Li, Z.-J. Zha, and M.~Wang, ``Dadnet: Dilated-attention-deformable
  convnet for crowd counting,'' in \emph{Proc. ACM MM}, 2019, pp. 1823--1832.

\bibitem{guo2019hierarchical}
D.~Guo, W.~Zhou, A.~Li, H.~Li, and M.~Wang, ``Hierarchical recurrent deep
  fusion using adaptive clip summarization for sign language translation,''
  \emph{IEEE Transactions on Image Processing}, vol.~29, pp. 1575--1590, 2019.

\bibitem{chen2018deepphys}
W.~Chen and D.~McDuff, ``Deepphys: Video-based physiological measurement using
  convolutional attention networks,'' in \emph{Proc. ECCV}, 2018, pp. 349--365.

\bibitem{vspetlik2018visual}
R.~{\v{S}}petl{\'\i}k, V.~Franc, and J.~Matas, ``Visual heart rate estimation
  with convolutional neural network,'' in \emph{Proc. BMVC}, 2018, pp. 3--6.

\bibitem{lee2020meta}
E.~Lee, E.~Chen, and C.-Y. Lee, ``Meta-rppg: Remote heart rate estimation using
  a transductive meta-learner,'' in \emph{Proc. ECCV}, 2020, pp. 392--409.

\bibitem{lu2021dual}
H.~Lu, H.~Han, and S.~K. Zhou, ``Dual-gan: Joint bvp and noise modeling for
  remote physiological measurement,'' in \emph{Proc. CVPR}, 2021, pp.
  12\,404--12\,413.

\bibitem{niu2020video}
X.~Niu, Z.~Yu, H.~Han, X.~Li, S.~Shan, and G.~Zhao, ``Video-based remote
  physiological measurement via cross-verified feature disentangling,'' in
  \emph{Proc. ECCV}, 2020, pp. 295--310.

\bibitem{gideon2021way}
J.~Gideon and S.~Stent, ``The way to my heart is through contrastive learning:
  Remote photoplethysmography from unlabelled video,'' in \emph{Proc. ICCV},
  2021, pp. 3995--4004.

\bibitem{wang2022synthetic}
Z.~Wang, Y.~Ba, P.~Chari, O.~D. Bozkurt, G.~Brown, P.~Patwa, N.~Vaddi,
  L.~Jalilian, and A.~Kadambi, ``Synthetic generation of face videos with
  plethysmograph physiology,'' in \emph{Proc. CVPR}, 2022, pp.
  20\,587--20\,596.

\bibitem{hsu2017deep}
G.-S. Hsu, A.~Ambikapathi, and M.-S. Chen, ``Deep learning with time-frequency
  representation for pulse estimation from facial videos,'' in \emph{Proc.
  IJCB}, 2017, pp. 383--389.

\bibitem{niu2018vipl}
X.~Niu, H.~Han, S.~Shan, and X.~Chen, ``Vipl-hr: A multi-modal database for
  pulse estimation from less-constrained face video,'' in \emph{Proc. ACCV},
  2018, pp. 562--576.

\bibitem{liu2020multi}
X.~Liu, J.~Fromm, S.~Patel, and D.~McDuff, ``Multi-task temporal shift
  attention networks for on-device contactless vitals measurement,'' in
  \emph{Proc. NeurIPS}, vol.~33, 2020, pp. 19\,400--19\,411.

\bibitem{niu2019robust}
X.~Niu, X.~Zhao, H.~Han, A.~Das, A.~Dantcheva, S.~Shan, and X.~Chen, ``Robust
  remote heart rate estimation from face utilizing spatial-temporal
  attention,'' in \emph{Proc. FG}, 2019, pp. 1--8.

\bibitem{vaswani2017attention}
A.~Vaswani, N.~Shazeer, N.~Parmar, J.~Uszkoreit, L.~Jones, A.~N. Gomez,
  {\L}.~Kaiser, and I.~Polosukhin, ``Attention is all you need,'' in
  \emph{Proc. NeurIPS}, 2017, pp. 5998--6008.

\bibitem{dosovitskiy2020image}
A.~Dosovitskiy, L.~Beyer, A.~Kolesnikov, D.~Weissenborn, X.~Zhai,
  T.~Unterthiner, M.~Dehghani, M.~Minderer, G.~Heigold, S.~Gelly \emph{et~al.},
  ``An image is worth 16x16 words: Transformers for image recognition at
  scale,'' in \emph{Proc. ICLR}, 2020, pp. 1--21.

\bibitem{liu2021efficientphys}
X.~Liu, B.~L. Hill, Z.~Jiang, S.~Patel, and D.~McDuff, ``Efficientphys:
  Enabling simple, fast and accurate camera-based vitals measurement,''
  \emph{arXiv:2110.04447}, 2021.

\bibitem{carionend}
N.~Carion, F.~Massa, G.~Synnaeve, N.~Usunier, A.~Kirillov, and S.~Zagoruyko,
  ``End-to-end object detection with transformers,'' in \emph{Proc. ECCV},
  2020, pp. 213--229.

\bibitem{bobbia2019unsupervised}
S.~Bobbia, R.~Macwan, Y.~Benezeth, A.~Mansouri, and J.~Dubois, ``Unsupervised
  skin tissue segmentation for remote photoplethysmography,'' \emph{Pattern
  Recognition Letters}, vol. 124, pp. 82--90, 2019.

\bibitem{stricker2014non}
R.~Stricker, S.~M{\"u}ller, and H.-M. Gross, ``Non-contact video-based pulse
  rate measurement on a mobile service robot,'' in \emph{Proc. IISRHIC}, 2014,
  pp. 1056--1062.

\bibitem{wieringa2005contactless}
F.~P. Wieringa, F.~Mastik, and A.~F. van~der Steen, ``Contactless multiple
  wavelength photoplethysmographic imaging: a first step toward “spo2
  camera” technology,'' \emph{Annals of Biomedical Engineering}, vol.~33,
  no.~8, pp. 1034--1041, 2005.

\bibitem{wang2014exploiting}
W.~Wang, S.~Stuijk, and G.~De~Haan, ``Exploiting spatial redundancy of image
  sensor for motion robust rppg,'' \emph{IEEE Transactions on Biomedical
  Engineering}, vol.~62, no.~2, pp. 415--425, 2014.

\bibitem{yu2019remote1}
Z.~Yu, X.~Li, and G.~Zhao, ``Remote photoplethysmograph signal measurement from
  facial videos using spatio-temporal networks,'' in \emph{Proc. BMVC}, 2019,
  pp. 1--12.

\bibitem{nowara2021benefit}
E.~M. Nowara, D.~McDuff, and A.~Veeraraghavan, ``The benefit of distraction:
  Denoising camera-based physiological measurements using inverse attention,''
  in \emph{Proc. ICCV}, 2021, pp. 4955--4964.

\bibitem{niu2017continuous}
X.~Niu, H.~Han, S.~Shan, and X.~Chen, ``Continuous heart rate measurement from
  face: A robust rppg approach with distribution learning,'' in \emph{Proc.
  IJCB}, 2017, pp. 642--650.

\bibitem{guohierarchical}
D.~Guo, W.~Zhou, H.~Li, and M.~Wang, ``Hierarchical lstm for sign language
  translation,'' in \emph{Proc. AAAI}, vol.~32, no.~1, 2018.

\bibitem{guoconnectionist}
D.~Guo, S.~Tang, and M.~Wang, ``Connectionist temporal modeling of video and
  language: a joint model for translation and sign labeling,'' in \emph{Proc.
  IJCAI}, 2019, pp. 751--757.

\bibitem{simonyan2014very}
K.~Simonyan and A.~Zisserman, ``Very deep convolutional networks for
  large-scale image recognition,'' in \emph{Proc. ICLR}, 2015, pp. 1--14.

\bibitem{tsou2020siamese}
Y.-Y. Tsou, Y.-A. Lee, C.-T. Hsu, and S.-H. Chang, ``Siamese-rppg network:
  Remote photoplethysmography signal estimation from face videos,'' in
  \emph{Proc. ACM SAC}, 2020, pp. 2066--2073.

\bibitem{song2021pulsegan}
R.~Song, H.~Chen, J.~Cheng, C.~Li, Y.~Liu, and X.~Chen, ``Pulsegan: Learning to
  generate realistic pulse waveforms in remote photoplethysmography,''
  \emph{IEEE Journal of Biomedical and Health Informatics}, vol.~25, no.~5, pp.
  1373--1384, 2021.

\bibitem{suncontrast}
Z.~Sun and X.~Li, ``Contrast-phys: Unsupervised video-based remote
  physiological measurement via spatiotemporal contrast,'' in \emph{Proc.
  ECCV}, 2022, pp. 492--510.

\bibitem{luneuron}
H.~Lu, Z.~Yu, X.~Niu, and Y.-C. Chen, ``Neuron structure modeling for
  generalizable remote physiological measurement,'' in \emph{Proc. CVPR}, 2023,
  pp. 18\,589--18\,599.

\bibitem{lin2019tsm}
J.~Lin, C.~Gan, and S.~Han, ``Tsm: Temporal shift module for efficient video
  understanding,'' in \emph{Proc. ICCV}, 2019, pp. 7083--7093.

\bibitem{liu2021swin}
Z.~Liu, Y.~Lin, Y.~Cao, H.~Hu, Y.~Wei, Z.~Zhang, S.~Lin, and B.~Guo, ``Swin
  transformer: Hierarchical vision transformer using shifted windows,'' in
  \emph{Proc. ICCV}, 2021, pp. 10\,012--10\,022.

\bibitem{yu2021transrppg}
Z.~Yu, X.~Li, P.~Wang, and G.~Zhao, ``Transrppg: Remote photoplethysmography
  transformer for 3d mask face presentation attack detection,'' \emph{IEEE
  Signal Processing Letters}, vol.~28, pp. 1290--1294, 2021.

\bibitem{baltrusaitis2018openface}
T.~Baltrusaitis, A.~Zadeh, Y.~C. Lim, and L.-P. Morency, ``Openface 2.0: Facial
  behavior analysis toolkit,'' in \emph{Proc. FG}, 2018, pp. 59--66.

\bibitem{yu2020autohr}
Z.~Yu, X.~Li, X.~Niu, J.~Shi, and G.~Zhao, ``Autohr: A strong end-to-end
  baseline for remote heart rate measurement with neural searching,''
  \emph{IEEE Signal Processing Letters}, vol.~27, pp. 1245--1249, 2020.

\bibitem{de2014improved}
G.~De~Haan and A.~Van~Leest, ``Improved motion robustness of remote-ppg by
  using the blood volume pulse signature,'' \emph{Physiological Measurement},
  vol.~35, no.~9, pp. 1913--1913, 2014.

\bibitem{hu2021rppg}
M.~Hu, D.~Guo, M.~Jiang, F.~Qian, X.~Wang, and F.~Ren, ``rppg-based heart rate
  estimation using spatial-temporal attention network,'' \emph{IEEE
  Transactions on Cognitive and Developmental Systems}, 2021.

\bibitem{niu2018synrhythm}
X.~Niu, H.~Han, S.~Shan, and X.~Chen, ``Synrhythm: Learning a deep heart rate
  estimator from general to specific,'' in \emph{Proc. ICPR}, 2018, pp.
  3580--3585.

\bibitem{macwan2019heart}
R.~Macwan, Y.~Benezeth, and A.~Mansouri, ``Heart rate estimation using remote
  photoplethysmography with multi-objective optimization,'' \emph{Biomedical
  Signal Processing and Control}, vol.~49, pp. 24--33, 2019.

\bibitem{carreira2017quo}
J.~Carreira and A.~Zisserman, ``Quo vadis, action recognition? a new model and
  the kinetics dataset,'' in \emph{Proc. CVPR}, 2017, pp. 6299--6308.

\bibitem{yang2016motion}
Y.~Yang, C.~Liu, H.~Yu, D.~Shao, F.~Tsow, and N.~Tao, ``Motion robust remote
  photoplethysmography in cielab color space,'' \emph{Journal of Biomedical
  Optics}, vol.~21, no.~11, pp. 117\,001--117\,001, 2016.

\bibitem{van2008visualizing}
L.~Van~der Maaten and G.~Hinton, ``Visualizing data using t-sne.''
  \emph{Journal of Machine Learning Research}, vol.~9, no.~11, pp. 2579--2605,
  2008.

\end{thebibliography}

\end{document}